%% file: main.tex
\documentclass[10pt,twocolumn,letterpaper]{article}
\usepackage[accsupp]{axessibility}  
\usepackage{cvpr}              
\usepackage{placeins}

\input{preamble}

\definecolor{cvprblue}{rgb}{0.21,0.49,0.74}
\usepackage[pagebackref,breaklinks,colorlinks,allcolors=cvprblue]{hyperref}

\def\paperID{33301} 
\def\confName{CVPR}
\def\confYear{2026}

\title{AffordGrasp: Cross-Modal Diffusion for Affordance-Aware Grasp Synthesis}

\author{
Xiaofei Wu$^1$, Yi Zhang$^1$, Yumeng Liu$^{3}$, Yuexin Ma$^1$, Yujiao Shi$^{1*}$, Xuming He$^{1,2*}$\\
$^1$ShanghaiTech University, Shanghai, China \\
$^2$Shanghai Engineering Research Center of Intelligent Vision and Imaging \\
$^3$University of Science and Technology of China
}

\begin{document}

\twocolumn[{
\renewcommand\twocolumn[1][]{#1}%
\maketitle

\begin{center}
\setlength{\abovecaptionskip}{0pt}
\setlength{\belowcaptionskip}{0pt}
\setlength{\tabcolsep}{2pt}
\vspace{-7ex}
\captionsetup{type=figure}
\begin{tabular}{ccc}
  \includegraphics[width=0.25\textwidth]{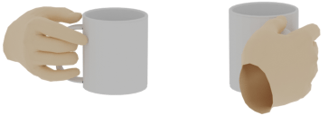} & \hspace{0.5cm} 
  \includegraphics[width=0.25\textwidth]{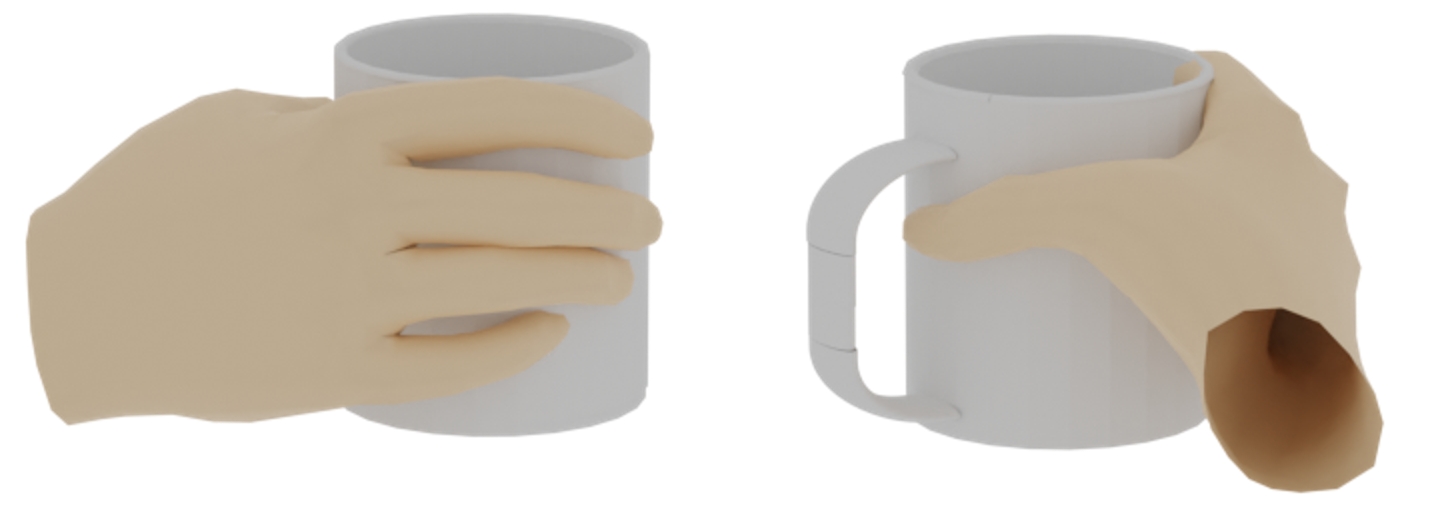} & \hspace{0.5cm}
  \includegraphics[width=0.2\textwidth]{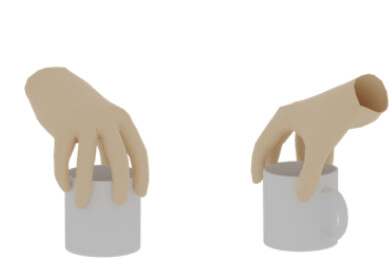} \\
  \small Grip the handle of the mug. & \hspace{0.5cm} 
  \small Wrap your hand around the mug. & \hspace{0.5cm} 
  \small Lift the mug to avoid spills. \\
\end{tabular}
\vspace{0ex}
\begin{tabular}{ccc}
  \includegraphics[width=0.25\textwidth]{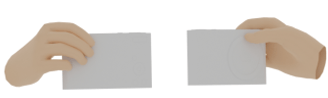} & \hspace{0.5cm}
  \includegraphics[width=0.25\textwidth]{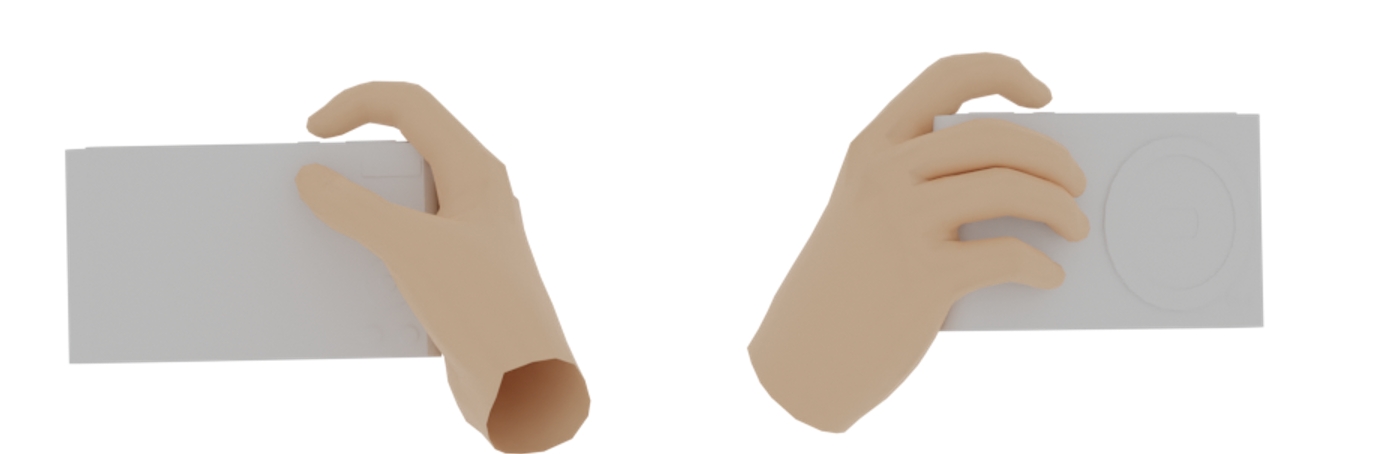} & \hspace{0.5cm}
  \includegraphics[width=0.25\textwidth]{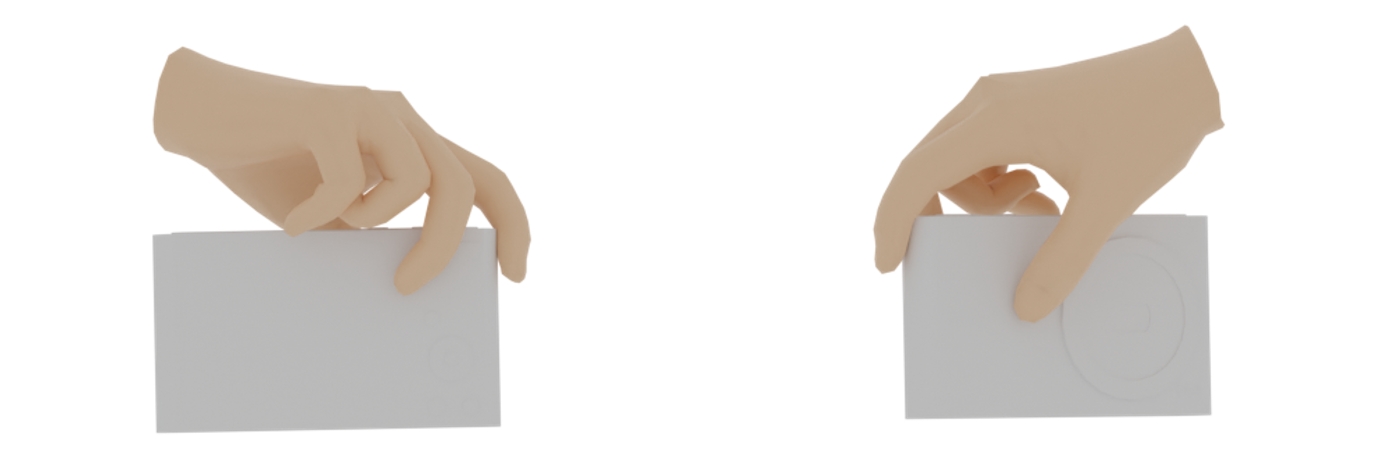} \\ 
  \hspace{0.8cm}\small Wrap your hand around the camera. & \hspace{0.5cm}
  \small Press the camera to take a photo. & \hspace{0.5cm}
  \small Wrap your palm around the camera. \\
\end{tabular}  
\vspace{-2ex}
\begin{tabular}{ccc}
  \hspace{0.5cm}\includegraphics[width=0.25\textwidth]{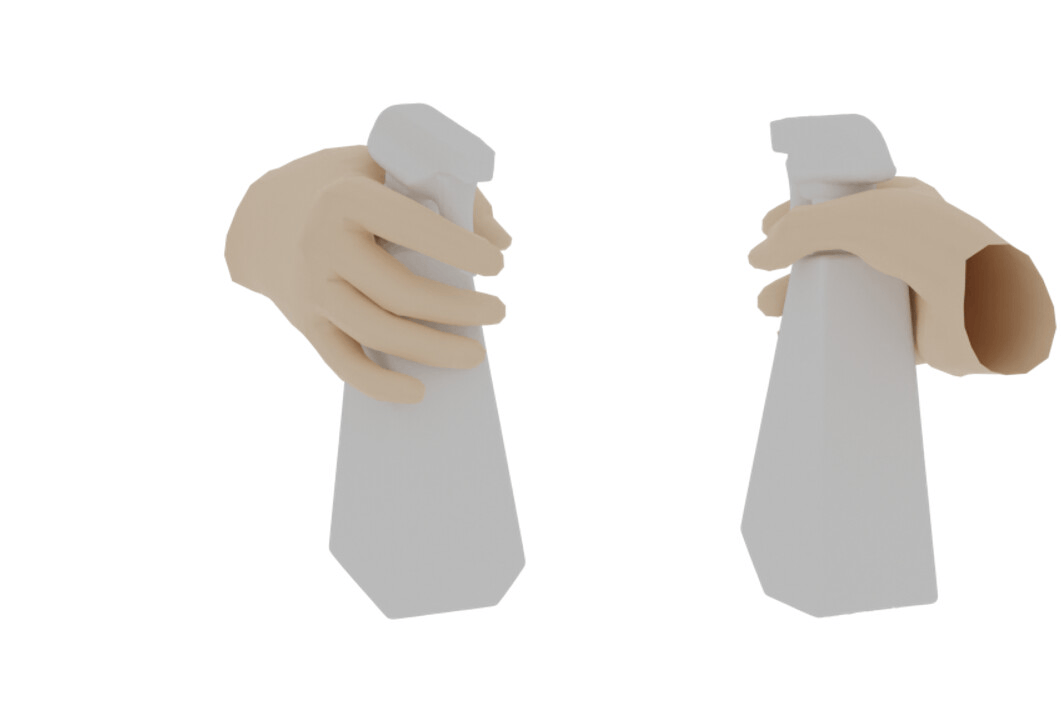} & \hspace{0.8cm}
  \includegraphics[width=0.25\textwidth]{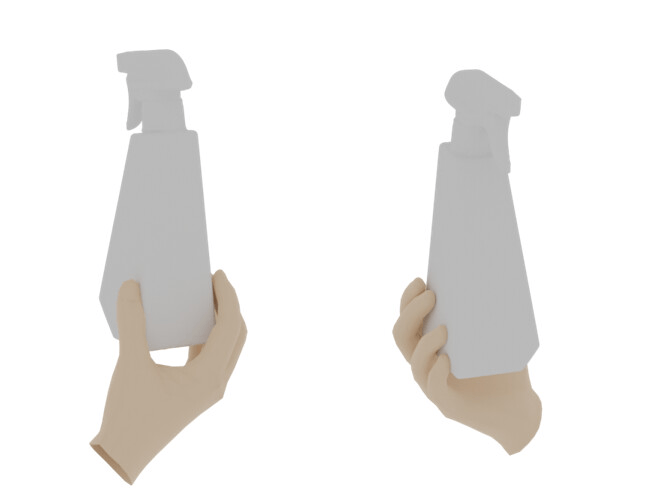} & \hspace{-1cm}
  \includegraphics[width=0.25\textwidth]{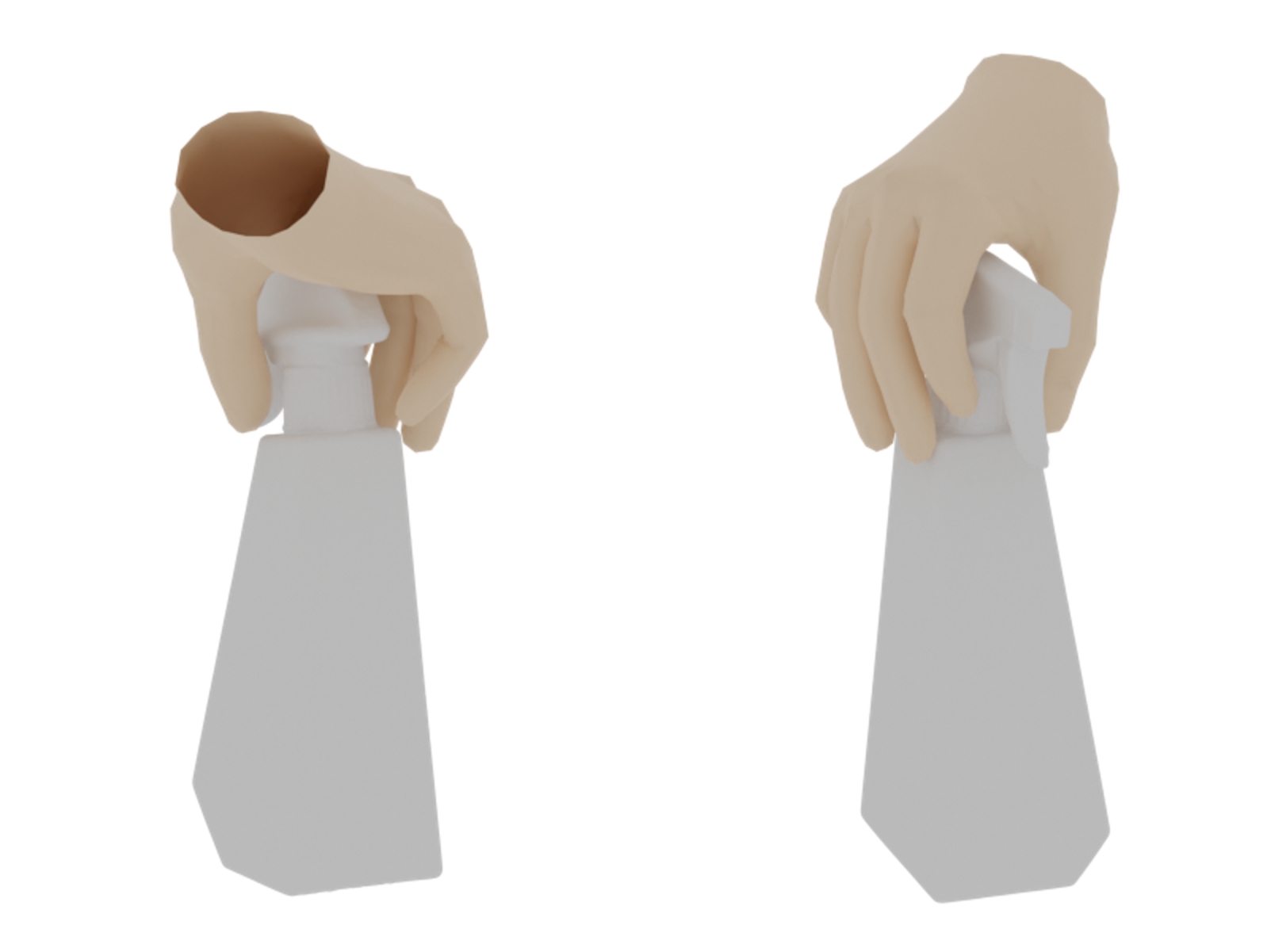} \\
   \hspace{2cm}\small Press the dispenser to pour. & \hspace{1.2cm}
  \small Support the dispenser from underneath. & \hspace{0.5cm}
  \small Twist the top of the dispenser to open it. \\
\end{tabular}
\vspace{3ex}
\caption{
\textbf{AffordGrasp} produces realistic and instruction-aligned hand grasps directly from textual descriptions.
For each object, we demonstrate the model’s capacity to generate semantically diverse and physically plausible grasps across three textual instructions, visualized from two viewpoints to reveal the precision and robustness of the synthesized hand poses.
}
\label{fig:teaser}
\end{center}

}]
\renewcommand{\thefootnote}{*}
\footnotetext{Corresponding authors.}
\input{sec/0_abstract}    
\input{sec/1_intro}

\input{sec/2_relat}
\input{sec/3_approach_AffordGrasp}

\input{sec/3_approach}
\input{sec/4_exp}
\input{sec/5_conclusion}
{
    \small
    \bibliographystyle{ieeenat_fullname}
    \bibliography{main}
}
\clearpage
\clearpage
\input{sec/X_suppl}
\end{document}

%% file: preamble.tex
\definecolor{myred}{RGB}{255, 99, 71}   
\definecolor{mygreen}{RGB}{60, 179, 113} 
\definecolor{myblue}{RGB}{30, 144, 255}

%% file: sec/0_abstract.tex
\begin{abstract}

Generating human grasping poses that accurately reflect both object geometry and user-specified interaction semantics is essential for natural hand–object interactions in AR/VR and embodied AI. However, existing semantic grasping approaches struggle with the large modality gap between 3D object representations and textual instructions, and often lack explicit spatial or semantic constraints, leading to physically invalid or semantically inconsistent grasps. In this work, we present AffordGrasp, a diffusion-based framework that produces physically stable and semantically faithful human grasps with high precision. We first introduce a scalable annotation pipeline that automatically enriches hand–object interaction datasets with fine-grained structured language labels capturing interaction intent. Building upon these annotations, AffordGrasp integrates an affordance-aware latent representation of hand poses with a dual-conditioning diffusion process, enabling the model to jointly reason over object geometry, spatial affordances, and instruction semantics. A distribution adjustment module further enforces physical contact consistency and semantic alignment. We evaluate AffordGrasp across four instruction-augmented benchmarks derived from HO-3D, OakInk, GRAB, and AffordPose, and observe substantial improvements over state-of-the-art methods in grasp quality, semantic accuracy, and diversity.  \href{https://wuxiaofei01.github.io/AffordGrasp_page}{Project Website: AffordGrasp}

\end{abstract}

%% file: sec/1_intro.tex
\section{Introduction}

Semantic-based grasp generation aims to synthesize human hand poses that interact with objects according to user instructions, enabling natural and intuitive interactions for AR/VR and robotic systems. However, traditional grasp generation approaches~\cite{Wu2024FastGraspEG,liu2023contactgen,jiang2021graspTTA,halo,grabdataset} rely solely on 3D object geometry and thus fail to reflect the user’s intended interaction. For instance, grasping a teacup by its rim versus holding its handle requires distinct semantic intent despite identical geometry. This underscores the need to jointly model object shape, linguistic intent, and interaction context to generate meaningful and physically valid grasps.

Recent semantic grasping frameworks combine point-cloud embeddings with textual instructions to condition diffusion models~\cite{Li2024SemGraspSG,Chang2024Text2GraspGS}. While effective to a degree, these models still struggle to produce high-precision grasps due to two key challenges:
(1) the substantial modality gap between raw 3D geometry and natural language makes direct fusion insufficient for fine-grained geometric–semantic alignment (e.g, distinguishing “grasp the handle” from “hold the rim”); and
(2) current diffusion pipelines lack explicit spatial and instruction-driven constraints, often yielding semantically incompatible contacts or physically unrealistic poses.
Although VLM-based annotation pipelines~\cite{Li2024SemGraspSG,Liu2024RealDexTH} seek to enhance semantic grounding through multi-turn question answering, these procedures remain susceptible to inconsistency and reduced controllability arising from error propagation across multiple reasoning steps, divergent reasoning paths, and context dependencies.

To address these challenges, we propose a method that directly generates fine-grained, structured language annotations for scalable and consistent dataset enrichment. Specifically, we automatically augment existing public hand–object interaction datasets with language labels that explicitly capture interaction intent. Building on this enriched data, we introduce \textit{AffordGrasp}, an efficient cross-modal generative framework designed to synthesize diverse grasp poses that satisfy both physical constraints and textual semantic instructions.

Our approach leverages a latent diffusion model~\cite{von-platen-etal-2022-diffusers} augmented with an affordance-aware representation of hand poses within a compact latent space. The diffusion process employs a dual-conditioning mechanism that systematically integrates physical plausibility and semantic guidance, effectively modeling the conditional distribution of hand poses given object properties and instructional prompts. Concretely, \textit{AffordGrasp} first generates local spatial cues aligned with instructions through an Affordance Generator and learns a low-dimensional latent representation of hand posture parameters using a Variational AutoEncoder. To ensure consistency with physical contact and semantic intent, we introduce a Distribution Adjustment Module that refines the latent representation during sampling based on object contact constraints and instruction semantics.

To validate our method, we construct four benchmarks based on HO-3D~\cite{ho3d}, OakInk~\cite{oakink}, GRAB~\cite{grabdataset}, and AffordPose~\cite{AffordPose}, each enhanced with fine-grained textual instructions describing specific grasp poses and object interactions. Extensive experiments demonstrate that \textit{AffordGrasp} significantly outperforms state-of-the-art baselines across all evaluation metrics, establishing a robust framework for advancing human grasp synthesis and embodied intelligence research.

In summary, our contributions are as follows:
\begin{itemize}
\item We introduce \textit{AffordGrasp}, a diffusion-based framework that generates physically stable and semantically meaningful grasps with high precision, without requiring test-time adaptation.
\item We propose the use of object affordance as complementary guidance for cross-modal fusion, bridging linguistic semantics and geometric representations to improve grasp intention understanding.
\item We develop a distribution adjustment module that maintains diffusion sampling stability while enforcing strict physical and semantic constraints on grasp poses.
\item Our method establishes a new state-of-the-art performance across multiple benchmarks through comprehensive quantitative and qualitative evaluations.
\end{itemize}

%% file: sec/2_relat.tex
\section{Related Work}
\paragraph{Grasp Synthesis.}
Grasp synthesis is critical for robot manipulation, animation, and human motion analysis~\cite{Liu2024RealDexTH, Zhang2024GraspXLGG}. 
We address the challenge of synthesizing realistic human grasps under two key constraints: semantic alignment with object functionality and physical plausibility. Existing methods typically predict MANO parameters~\cite{grabdataset, oakink, Wu2024FastGraspEG} or joint positions~\cite{halo} using generative models, while Liu \textit{et al.}\cite{Li2024SemGraspSG, Chang2024Text2GraspGS} suffer from a modality gap between 3D geometry and language, weakening the alignment with semantic intent and reducing attention to critical affordances.To overcome this, we introduce object affordances as cross-modal cues that bridge geometry and language. By leveraging affordance-aware features, our architecture improves semantic grounding while preserving fine-grained hand-object geometric fidelity.
\begin{figure*}[t]
    \centering
    \includegraphics[width=\textwidth]{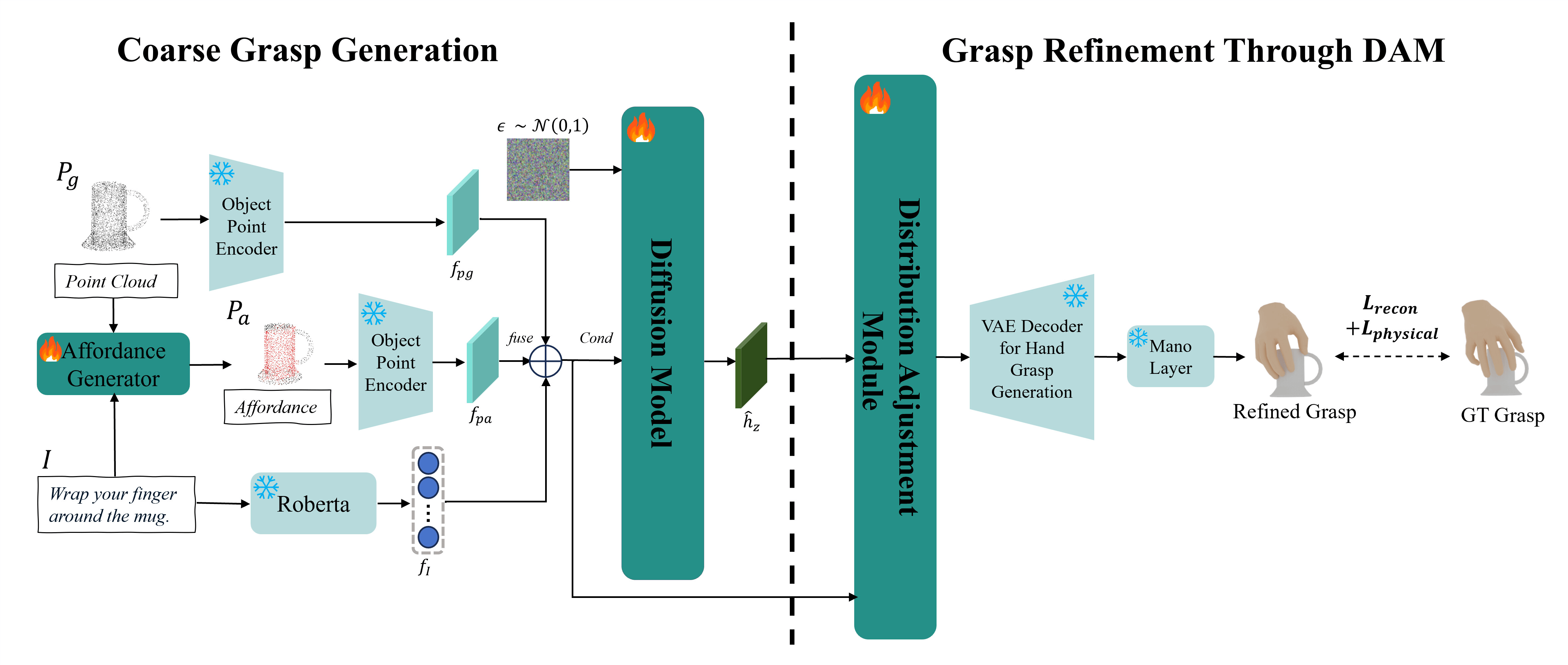}
    \caption{\textbf{Overview of AffordGrasp.}We integrate language instructions with object point clouds and employ the Affordance Generator to predict point-wise confidence features, which are aggregated into the final affordance map to enhance spatial detail and align linguistic semantics with 3D structures.The right part employs a DAM module, which ensures that the synthesized grasping poses generated by the LDM model align with physical constraints and language semantics.} 
    \label{fig:overview}
\end{figure*}

\paragraph{Affordance in Hand-Object Interaction.} 
Understanding hand-object interaction is critical for applications in AR, VR, and embodied AI. While existing methods emphasize contact cues to ensure physical plausibility and detailed modeling~\cite{brahmbhatt2019contactgrasp,jiang2021graspTTA,Liu2024EasyHOIUT,Wei2024GraspAY,Ye2024GHOPGH,corona2020ganhand}, they often overlook object functionality and affordances. Recent efforts provide object affordance annotations~\cite{oakink,AffordPose}, but focus mainly on geometric properties, lacking interaction-centric semantic reasoning essential for task-oriented grasping. SemGrasp~\cite{Li2024SemGraspSG} leverages multi-modal language models to unify object, grasp, and language representations, but relies on 2D projections of 3D data, which introduces occlusions and limits instruction fidelity.
In this work, we propose a 3D-native, interaction-aware framework that directly utilizes 3D spatial features to predict affordance categories. These features are integrated with a language model to generate precise, context-aware task instructions. Our method bridges spatial geometry and semantic understanding, enabling more accurate and robust task-driven grasp synthesis.

\paragraph{Denoising Diffusion Probabilistic Models.}
Denoising diffusion models~\cite{ddim, stablediffusion, Liu2022CompositionalVG, Kwon2022DiffusionMA,wu2025pack} generate data by reversing a structured noise corruption process through a learned forward-backward stochastic framework. Recent methods incorporate gradient-based objectives during sampling to steer generation~\cite{Wu2024AffordDPGD, Yang2024GuidanceWS}, but often struggle under high-noise conditions. In particular, aggressive gradient updates in early diffusion steps can push samples off the learned data manifold, resulting in irreversible distribution shifts and degraded output quality.

%% file: sec/3_approach_AffordGrasp.tex
\section{Approach - AffordGrasp}

We present AffordGrasp, a diffusion-based framework for generating human grasp poses that are both physically plausible and semantically aligned with user instructions. Given an object point cloud $P_g$ and a textual instruction $I$, the goal is to generate a functional grasp pose $h_p$, represented by MANO parameters~\cite{MANO:SIGGRAPHASIA:2017}. 

AffordGrasp explicitly models the interplay between instruction semantics, object geometry, and grasp intent through three integrated components: (1) an \textbf{Affordance Generator} that predicts interaction-relevant object regions from language–geometry inputs, (2) a \textbf{Cross-Modal Diffusion Model} that synthesizes grasp poses conditioned on multi-modal cues, and (3) a \textbf{Distribution Adjustment Module} (DAM) that refines the denoised latent representation to enhance contact realism and semantic precision.

\subsection{Affordance Generator}
\label{subsec: affordance}
Semantic grasp generation requires aligning the user instruction with geometric cues on the object. We therefore train an Affordance Generator that estimates point-wise affordance probabilities, indicating the relevance of each object point to the instruction. 
Given $(P_g, I)$, the generator predicts an affordance map $P_a$, highlighting a semantically grounded local region in the object point cloud. This region serves as an intermediate representation that explicitly links language semantics to a geometric structure, reducing the cross-modal difficulty faced by prior methods. 

Training the affordance generator is hindered by two main challenges: the lack of large-scale and diverse affordance datasets, and the severe imbalance between
affording and non-affording object points. Among existing datasets, only AffordPose~\cite{AffordPose} provides affordance annotations, while datasets such as GRAB~\cite{grabdataset} and OakInk~\cite{oakink} lack such labels. This limitation reduces the geometric diversity of objects within the training data.

To address the first challenge, we generate pseudo-labels using a self-looping annotation engine. Specifically, we adopt the network architecture proposed by LASO~\cite{laso} as our affordance generator and initially train it on the AffordPose~\cite{AffordPose} dataset. The trained model is then used in a self-training loop to annotate OakInk and GRAB, thereby expanding the dataset and enriching the geometric diversity of the training objects.

To address the class imbalance issue in affordance prediction, we employ a combined objective of Focal Loss~\cite{Lin2017FocalLF} and Dice Loss~\cite{Sudre2017GeneradlisedDO} to optimize the model:
\begin{equation}
\mathcal{L} = \mathcal{L}_{\text{focal}} + \lambda_1\mathcal{L}_{\text{dice}},
\end{equation}
where $\lambda_1$ is a balancing hyperparameter. Details of $\mathcal{L}_{\text{focal}}$ and $\mathcal{L}_{\text{dice}}$ are presented in the appendix Sec. 9.2. 

\subsection{Text and Affordance Guided Grasp Generation}

\label{subsec:diffusion}

Generating functional hand–object interactions requires jointly modeling semantic intent derived from language and geometric feasibility informed by object shape and affordance cues. To this end, we propose a cross-modal latent diffusion model conditioned on the triplet $\mathcal{C}=\{I, P_g, P_a\}$, which encodes the textual instruction, object geometry, and predicted affordance, respectively.

We first apply modality-specific encoders to extract feature representations from each input. Following LASO~\cite{laso}, the language instruction $I$ is encoded using RoBERTa~\cite{Liu2019RoBERTaAR}. The object point cloud $P_g$ and the affordance point cloud $P_a$ are processed by two independent PointNet encoders ($E_g,E_a$)~\cite{Qi2016PointNetDL}. The resulting features are then fused to construct a unified conditioning vector $f=\{{f_I,f_{pg},f_{pa}\}}$. The formal definition is:
\begin{equation}
    \begin{aligned}
        f_I=\text{RoBERTa}(I),f_{pg}&=E_g(P_g),f_{pa}=E_a(p_a)\\
    \end{aligned}
\end{equation}
To better preserve spatial structure in grasp poses, we encode the ground-truth hand mesh vertices $h_v^{gt} \in \mathbb{R}^{778 \times 3}$ into a compact latent representation  $h_z = \mathcal{E}(h_v^{gt})$ using a pre-trained  autoencoder~\cite{Wu2024FastGraspEG}. A conditional diffusion model~\cite{stablediffusion} is then trained to learn the distribution of latent hand embeddings conditioned on $f$. The forward diffusion process is defined as:
\begin{equation}
z^t = \sqrt{\alpha_t} z^0 + \sqrt{1 - \alpha_t} \epsilon,
\end{equation}
 where $\epsilon \sim \mathcal{N}(0, \mathbf{I})$ and $\alpha_t$ is a predefined noise schedule. At each timestep $t$, a noise prediction network $\epsilon_\theta$ estimates the additive Gaussian noise $\epsilon$. The decoder maps the denoised latent representation to MANO parameters $h_p \in \mathbb{R}^{61}$, which are further passed through a differentiable MANO layer to reconstruct the hand mesh $h_m$. The training objective of the diffusion module is defined as:
\begin{equation}
L_{LDM}:=\mathbb{E}_{\mathcal{E}(h_v),\epsilon\thicksim\mathcal{N}(0,1),t}\Big[\|\epsilon-\epsilon_\theta(z^t,f,t)\|_2^2\Big],
\label{eq:diff} 
\end{equation}
where $\epsilon_\theta(z^t, f, t)$ denotes the conditional denoising U-Net, and $z^t$ is the noisy latent obtained by perturbing the encoded hand representation $h_z$ with Gaussian noise at timestep $t$. Through iterative denoising, the model learns to reconstruct realistic and semantically consistent hand meshes $h_m$ from noise, guided by both linguistic and geometric cues.

\begin{figure}[t]
    \centering
    \includegraphics[width=0.35\textwidth]{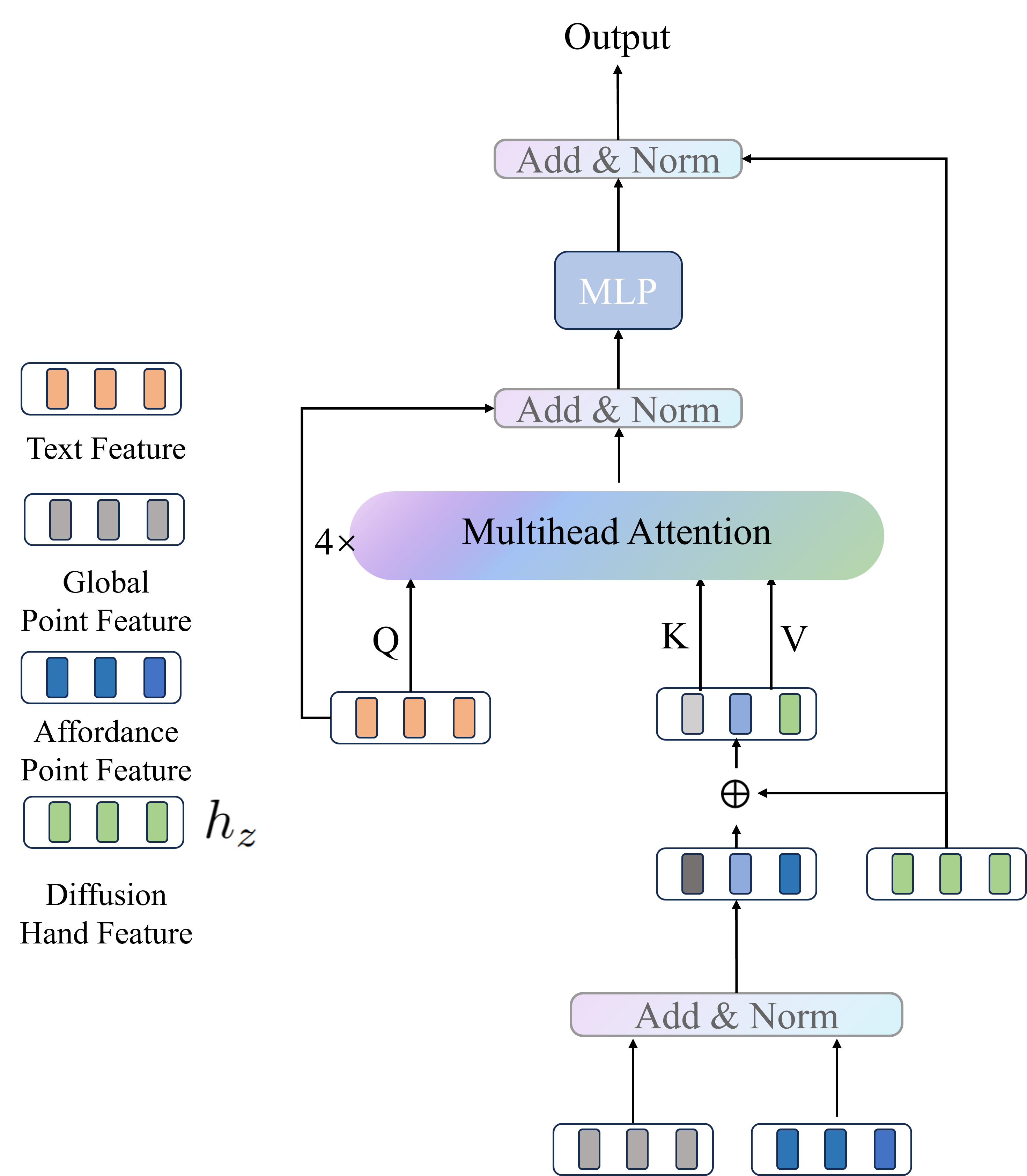}
    \vspace{-0.5em}
    \caption{\textbf{Distribution Adjustment Module (DAM) Architecture.} Hand and object features are fused and aligned with language instructions to produce stable, instruction-consistent grasps.
    }
    \label{fig:DAM}
\end{figure}

\subsection{Distribution Adjustment Module}
\label{subsec:DAM}

To generate more accurate and physically plausible grasping poses, we introduce the Distribution Adjustment Module (DAM). As illustrated in Fig.~\ref{fig:DAM}, DAM is a lightweight fusion framework that refines the latent grasp representation $\hat{h}_z$ predicted by the diffusion module. It integrates the conditioning feature $f$ with $\hat{h}_z$ to produce a refined latent $\tilde{h}_z$, ensuring the final pose aligns closely with both contextual and physical constraints.

According to Eq.~\ref{eq:diff}, the diffusion model $\epsilon_{\theta}$ predicts the noise term $\epsilon$ during training, which prevents its output from being directly used as input to the Distribution Adjustment Module (DAM). To address this issue, we propose a simple yet efficient approximation strategy that converts the noise prediction of the diffusion model into the corresponding latent hand pose representation. This process can be expressed as:
\begin{align}
\hat{h}_z = \frac{1}{\sqrt{\alpha_t}} (z^t - \sqrt{1-\alpha_t} \epsilon_{\theta}(z^t, f, t)).
\label{equ:approximation-1}
\end{align}

As shown in Fig.~\ref{fig:DAM}, we first construct the spatial representation $f_{\text{spatial}}$ by fusing the global point feature ($f_{pg}$) and the affordance point feature ($f_{pa}$) with the diffusion hand feature $\hat{h}_z$ (from Eq.~\ref{equ:approximation-1}). To effectively balance geometric detail and task intent, this representation interacts with the instruction embedding $f_I$ through a multi-head attention (MHA) module. This process employs a dual residual mechanism: the first residual connection preserves the instruction semantics ($f_I$) after the MHA, and the second preserves the original hand representation ($\hat{h}_z$) after the MLP module. This design contributes to improved performance while enhancing the network's expressive capacity. Unlike training-free methods~\cite{Wu2024AffordDPGD, Yang2024GuidanceWS} that increase inference time, our DAM is a lightweight, single-pass refinement module applied post-sampling, resulting in minimal inference overhead. The process is formulated as:
\begin{equation}
    \begin{aligned}
        f_{\text{spatial}} &= \operatorname{Norm}(f_{pg} + f_{pa}) + \hat{h}_z, \\
        f_{\text{align}} &= \operatorname{Attention}(f_I, f_{\text{spatial}}, f_{\text{spatial}}) + f_I, \\
         \tilde{h}_z &= \operatorname{Norm}(\operatorname{MLP}(f_{\text{align}}) + \hat{h}_z),
        \label{equ:dam-2}
    \end{aligned}
\end{equation}
where $f_{pg}$, $f_{pa}$, and $f_I$ denote the embeddings extracted from $P_g$, $P_a$, and $I$ using pretrained models. According to Eq.~\ref{equ:dam-2}, the proposed DAM module integrates the latent hand feature with both the instruction and the object geometry, thereby improving physical feasibility and instruction adherence.

During DAM training, the diffusion model $\epsilon_{\theta}$ remains frozen. The predicted latent representation $\hat{h}_z$ from Eq.~\ref{equ:approximation-1} is used as the input to the DAM module. Since $\hat{h}_z$ may contain inaccuracies, the DAM further refines it to produce a more reliable grasp latent as below:
\begin{equation}
h_m = \text{MANO(Decoder}(\tilde{h}_z)), \quad \tilde{h}_z=\mathrm{DAM}(\hat{h}_z,f).
\label{equ:dam}
\end{equation}

The overall training objective, detailed in Eq.~\ref{Loss:total}, combines reconstruction and physical constraint losses following prior works~\cite{Wu2024FastGraspEG, jiang2021graspTTA}:
\begin{align}
\mathcal{L} = &\mathcal{L}_{\mathrm{recon}}(h_{v},\, h_{p},\, h_v^{\mathrm{gt}},\, h_p^{\mathrm{gt}}) \nonumber \\ 
&+  \lambda_2  \mathcal{L}_{\mathrm{physical}}(h_{m},\, h_m^{\mathrm{gt}},\, P_g),
\label{Loss:total}
\end{align}
where $\lambda_{2}$ is a balancing coefficient. The Decoder reconstructs the refined latent code $\tilde{h}_z$ into hand parameters $h_p$, which are then passed through the MANO layer to produce the hand mesh $h_m$. The loss $\mathcal{L}_{\mathrm{recon}}$ supervises the reconstruction of the hand pose by encouraging the predicted MANO parameters, vertices to align with their corresponding ground truth. In contrast, $\mathcal{L}_{\mathrm{physical}}$ enforces physically plausible hand–object interactions by penalizing object–object interpenetration and promoting consistent contact patterns.

By jointly optimizing reconstruction accuracy and physical constraints, the DAM learns to refine the diffusion model’s latent predictions, yielding final grasp poses $h_m$ that are both task-compliant and physically plausible. Additional implementation details are provided in Appendix Sec. 11.

%% file: sec/3_approach.tex
\begin{table*}[t]
    \centering
    \setlength{\abovecaptionskip}{0pt}
    \setlength{\belowcaptionskip}{0pt}
    \resizebox{\linewidth}{!}{
    \begin{tabular}{c|c|ccc|ccc}
        \hline
        Dataset&
        Method &
        \begin{tabular}{c} Penetration \\ Volume $\downarrow$ \end{tabular} &
        \begin{tabular}{c} Simulation \\ Displacement $\downarrow$ \end{tabular} &
        \begin{tabular}{c} Contact \\ Ratio $\uparrow$ \end{tabular} &
        \begin{tabular}{c} Entropy $\uparrow$ \end{tabular} &
        \begin{tabular}{c} Cluster \\ Size $\uparrow$ \end{tabular} &
        \begin{tabular}{c} ACC $\uparrow$ \end{tabular} \\
        \hline
        OakInk~\cite{oakink}
            &TTA~\cite{jiang2021graspTTA} & 8.21& 2.33& 97& 2.82& 2.81& 60.83\%\\
            &FastGrasp~\cite{Wu2024FastGraspEG} & 7.88& 2.27& 88& 2.88& 3.42& 78.05\%\\
            &D-VQVAE~\cite{D-vqvae} & 7.33& 2.44& 91& 2.88& 3.57& 76.4\%\\
            &Ours(ControlNet) & 8.08& 3.11& 81& 2.87& 3.44& 76.94\%\\
            &\textbf{Ours} & \textbf{7.31}& \textbf{1.43}& \textbf{98}& \textbf{2.94}& \textbf{3.74}& \textbf{80.08\%}\\
        \hline
        GRAB~\cite{grabdataset}
        &TTA~\cite{jiang2021graspTTA} & 6.51& 1.22& 91& 2.73& 1.41& 55.00\%\\
        &FastGrasp~\cite{Wu2024FastGraspEG} & 4.61 & \textbf{1.20}& 94& 2.76 & 1.96 & 61.50\%\\
        &D-VQVAE~\cite{D-vqvae} & 8.04& 1.82& 89& 2.88& 3.41& 57.50\%\\
        &Ours(ControlNet) & 6.78& 1.57& 93 & 2.76& 3.47& 60.50\%\\
        &\textbf{Ours} & \textbf{3.06}& \textbf{1.66}& \textbf{94}& \textbf{2.91}& \textbf{3.53}& \textbf{62.50\%}\\
        \hline
    \end{tabular}
    }
\caption{Quantitative comparison on the \textbf{OakInk} and \textbf{GRAB} dataset. We compare our results with  baselines as well as with a framework where the DAM module is replaced by the ControLNet~\cite{controlnet} structure. Our method achieves the best performance on all evaluation metrics. }
    \label{tab:in-domain}
\end{table*}

\begin{table*}[t]
    \centering
    \setlength{\abovecaptionskip}{0pt}
    \setlength{\belowcaptionskip}{0pt}
    \resizebox{\linewidth}{!}{
    \begin{tabular}{c|c|ccc|ccc}
        \hline
        Dataset&
        Method &
        \begin{tabular}{c} Penetration \\ Volume $\downarrow$ \end{tabular} &
        \begin{tabular}{c} Simulation \\ Displacement $\downarrow$ \end{tabular} &
        \begin{tabular}{c} Contact \\ Ratio $\uparrow$ \end{tabular} &
        \begin{tabular}{c} Entropy $\uparrow$ \end{tabular} &
        \begin{tabular}{c} Cluster \\ Size $\uparrow$ \end{tabular} &
        \begin{tabular}{c} ACC $\uparrow$ \end{tabular} \\
        \hline
        HO-3D~\cite{ho3d}
            &TTA~\cite{jiang2021graspTTA} & 12.55& 3.22& 95 & 2.55& 2.99& 66\%\\
            &FastGrasp~\cite{Wu2024FastGraspEG} & 14.45& 2.73 & 96& 2.81& 2.23& 52.00\%\\
            &D-VQVAE~\cite{D-vqvae} & 13.12& \textbf{2.33}& 95& 2.78& 3.63& 64.00\%\\
            &Ours(ControlNet) & 16.06& 2.38& 97 & 2.84& 3.52& 51.00\% \\
            &\textbf{Ours} & \textbf{7.38}& \textbf{2.33}& \textbf{97}& \textbf{2.85}& \textbf{3.70}& \textbf{72.00\%}\\
        \hline
        AffordPose~\cite{AffordPose}
        &TTA~\cite{jiang2021graspTTA} & 19.41& 4.32& 91& 2.89& 2.97& 42.56\%\\
        &FastGrasp~\cite{Wu2024FastGraspEG} & 22.75& 3.77& 88 & 2.83& 3.77& 54.08\%\\
        &D-VQVAE~\cite{D-vqvae} & 21.43& 4.18& 91& 2.91 & 3.64 & 63.99\%\\
        &Ours(ControlNet) & 24.77& 4.78& 88& 2.88& 3.55& 52.45\%\\
        &\textbf{Ours} & \textbf{10.36}& \textbf{3.59}& \textbf{91}& \textbf{2.92} & \textbf{3.93}& \textbf{69.71\%}\\
        \hline
    \end{tabular}
    }
\caption{Comparison with previous methods on the \textbf{HO-3D} and \textbf{AffordPose} dataset, where our model is trained on the GRAB~\cite{grabdataset} dataset. Our model achieves state-of-the-art performance on two out-of-domain dataset, setting new benchmarks.}
\label{tab:out-of-domain}
\end{table*}

\vspace{2ex}

\begin{figure}[t]
    \centering
    \includegraphics[width=0.4\textwidth]{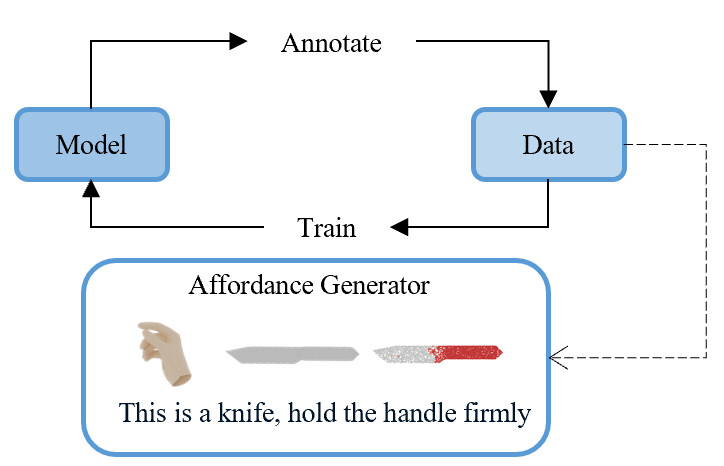}
    \vspace{-1em}
    \caption{\textbf{Affordance Annotation.} Implement an automated self-training pipeline that first assigns pseudo-labels to unlabeled data, then iteratively optimizes the model using these refined annotations.}
    \label{fig:data-engine}
\end{figure}
\begin{table*}[t]
    \centering
    \setlength{\abovecaptionskip}{0pt}
    \setlength{\belowcaptionskip}{0pt}
    \resizebox{\linewidth}{!}{
    \begin{tabular}{c|c|ccc|ccc}
        \hline
        Dataset&
        Method &
        \begin{tabular}{c} Penetration \\ Volume $\downarrow$ \end{tabular} &
        \begin{tabular}{c} Simulation \\ Displacement $\downarrow$ \end{tabular} &
        \begin{tabular}{c} Contact \\ Ratio $\uparrow$ \end{tabular} &
        \begin{tabular}{c} Entropy $\uparrow$ \end{tabular} &
        \begin{tabular}{c} Cluster \\ Size $\uparrow$ \end{tabular} &
        \begin{tabular}{c} ACC $\uparrow$ \end{tabular} \\
        \hline
        OakInk~\cite{oakink}
            &w/o object affordance & 8.27& \textbf{1.22}& 97 & 2.88 & 3.81 & 76.56\%\\
            &w/o DAM & 8.12& 1.77& 97 & 2.89& \textbf{4.07}& 79.11\%\\
            &\textbf{Whole pipeline} & \textbf{7.31}& 1.43& \textbf{98}& \textbf{2.94}& 3.74& \textbf{80.08\%}\\
        \hline
        GRAB~\cite{grabdataset}
        &w/o object affordance & 4.32& \textbf{1.55}& 92& 2.79& 3.61& 55.00\% \\
        &w/o DAM & 4.91& 1.71& 90& 2.84& \textbf{3.77}& \textbf{63.00\%}\\
        &\textbf{Whole pipeline} & \textbf{3.06}& 1.66& \textbf{94}& \textbf{2.91}& 3.53& 62.50\%\\
        \hline
        HO-3D~\cite{ho3d}
            &w/o object affordance & 8.88& \textbf{2.29}& 95& \textbf{2.85}& \textbf{3.71}& 71.00\%\\
            &w/o DAM & 11.21& 2.44& 91& 2.79& 3.64& 69.00\%\\
            &\textbf{Whole pipeline} & \textbf{7.38}& 2.33& \textbf{97}& \textbf{2.85}& 3.70& \textbf{72.00\%}\\
        \hline
        AffordPose~\cite{AffordPose}
        &w/o object affordance & 11.29& 3.78& 88& 2.84& 3.87& 67.81\%\\
        &w/o DAM & 12.32& 3.64& \textbf{91}& 2.88& \textbf{3.94}& 70.21\%\\
        &\textbf{Whole pipeline} & \textbf{10.36}& \textbf{3.59}& \textbf{91}& \textbf{2.92} & 3.83 & \textbf{72.43\%} \\
        \hline
    \end{tabular}
    }
\caption{Ablation study results on the \textbf{GRAB, OakInk, HO-3D, AffordPose} datasets~\cite{grabdataset,oakink,ho3d,AffordPose}. 
    The evaluation of the HO-3D and AffordPose is an out-of-domain generalization test, where the model is trained using the GRAB dataset.}
    \label{tab:ablation-study}
    \vspace{-2ex}
\end{table*}

\begin{figure*}[t]
\centering
\small  

\begin{minipage}{0.03\textwidth}
    \raggedleft
    \raisebox{1cm}{\rotatebox{90}{TTA}} \\
    \raisebox{1cm}{\rotatebox{90}{ControLNet}} \\
    \raisebox{1cm}{\rotatebox{90}{D-VQVAE}} \\ 
    \raisebox{1cm}{\rotatebox{90}{FastGrasp}} \\ 
    \raisebox{1cm}{\rotatebox{90}{\textbf{Ours}}}
\end{minipage}%
\begin{minipage}{0.4\textwidth}
    \centering
    \includegraphics[width=\textwidth]{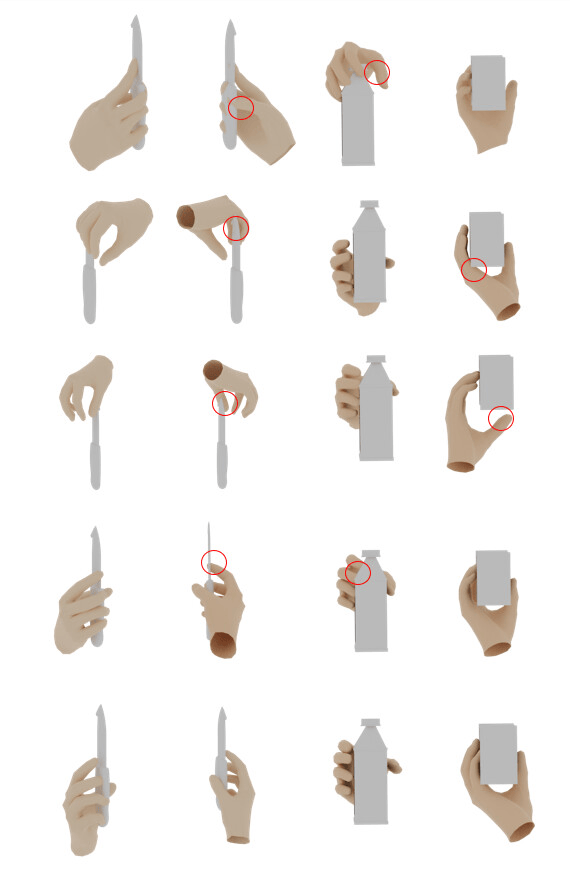}
    \vspace{-4ex}
    \captionsetup{labelformat=empty}
    \caption*{
        \centering
        OakInk~\cite{oakink}\\
        \textbf{Left:} This is a knife, hold the handle.\\
        \textbf{Right:} Wrap your hand around the bottle.
    }
    \vspace{-2ex}
\end{minipage}%
\hspace{0.005\textwidth}
\vrule width 0.5pt
\hspace{0.005\textwidth}
\begin{minipage}{0.4\textwidth}
    \centering
    \includegraphics[width=\textwidth]{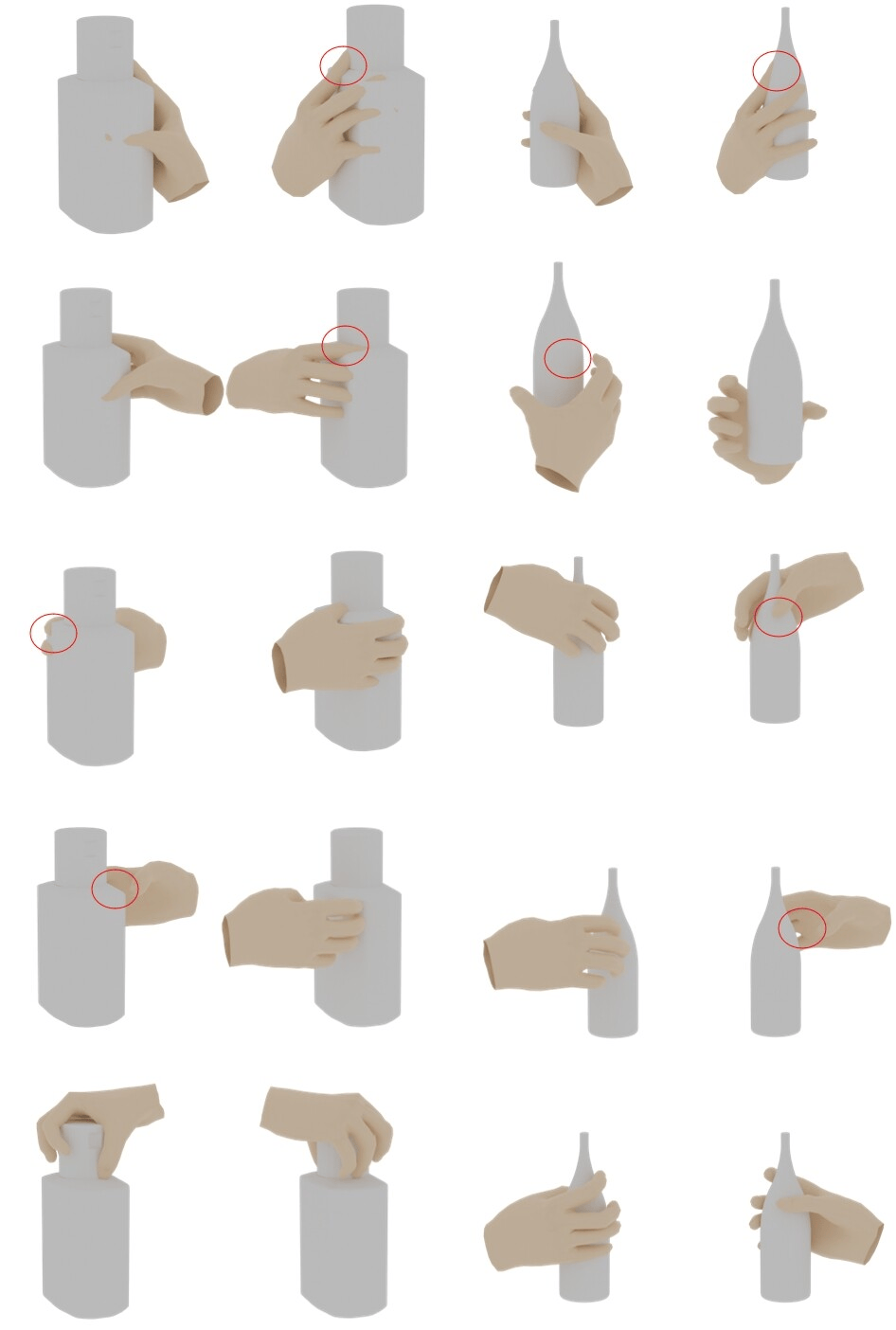}
    \vspace{-2ex}
    \captionsetup{labelformat=empty}
    \caption*{
        \centering
        AffordPose~\cite{AffordPose}\\
        \textbf{Left:} Twist the cap off the bottle.\\
        \textbf{Right:} Hold the bottle by wrapping your fingers.
    }
    \vspace{-2ex}
\end{minipage}

\caption{Qualitative comparisons with state-of-the-art methods on GRAB, OakInk datasets. Each pair (two columns) visualizes the generated grasps from two different views. Our method demonstrates a significant reduction in object penetration compared to other methods.}
\vspace{-4ex}
\label{fig:comparison-indomain}
\end{figure*}

\subsection{Inference}
\label{3.4}

    

Our inference process commences by sampling noise $\epsilon \sim \mathcal{N}(0,1)$ and condition it on the  $f$, which guides the diffusion model to synthesize coarse grasp poses in the latent space. To improve sampling efficiency, we adopt the DDIM framework~\cite{ddim}, governed by the following update rule:

\begin{equation}
    \begin{aligned}
        \boldsymbol{z}_{t-1} &= \sqrt{\alpha_{t-1}} \left( \frac{\boldsymbol{z}_t - \sqrt{1-\alpha_t} \epsilon_\theta(\boldsymbol{z}_t, f)}{\sqrt{\alpha_t}} \right) \\
        &\quad + \sqrt{1-\alpha_{t-1}-\sigma_t^2} \cdot \epsilon_\theta(\boldsymbol{z}_t, f) + \sigma_t \epsilon,
    \end{aligned}
\end{equation}
where $\alpha_t$ represents the DDIM scheduling parameters, and $\epsilon_\theta$ corresponds to the conditional denoising U-Net~\cite{von-platen-etal-2022-diffusers}.

To further strengthen the model’s conditional dependence on both physical constraints and linguistic semantics, we refine the generated sample using DAM. The refinement is defined as:
\begin{equation}
    \tilde{z}_1 = \mathrm{DAM}(z_1,f),
\end{equation}
where $\tilde{z}_1$ represents the optimized hand representation in the latent space, and $z_1$ denotes the DDIM output. The refined latent vector is then decoded into the hand mesh $h_m$ through a decoder and MANO layer. This process can be formally expressed as follows:
\begin{equation}
    h_m = \text{MANO(Decoder}(\tilde{z}_1)).
\end{equation}

Our framework generates hand mesh $h_m$ satisfies both linguistic plausibility and physical feasibility. In contrast to prior approaches that fuse the instruction and object point cloud to generate the final pose, our method mitigates cross-modal challenges. By narrowing the gap between language and geometric modalities, it produces hand poses that are more physically plausible and semantically aligned with the instructions.

%% file: sec/4_exp.tex



\section{Experiment}

\subsection{Automated Labeling for Dataset Enrichment}
\label{data-engine}
To enrich the OakInk~\cite{oakink} and GRAB~\cite{grabdataset} datasets—both centered on hand-object interactions—we introduce an automated pipeline for generating instruction annotations. Starting with the AffordPose~\cite{AffordPose} dataset, we perform cold-start training of a classifier~\cite{yu2021pointbert} to produce initial language labels, which are iteratively refined via error analysis to improve consistency. These refined annotations, combined with the initial dataset, support multiple rounds of training and labeling for full annotation coverage. Finally, we incorporate large language models~\cite{qwen2} to generate task-oriented, step-by-step instructional text, further enhancing semantic richness. Implementation details are in the Appendix.

We evaluate our method on four benchmarks: OakInk~\cite{oakink}, GRAB~\cite{grabdataset}, HO-3D~\cite{ho3d}, and AffordPose~\cite{AffordPose}, following standard experimental protocols. For in-domain evaluation, we train and test on OakInk and GRAB; the latter includes 51 objects grasped by 10 subjects, while OakInk offers a larger-scale dataset with 1,700 objects manipulated by 12 subjects. For cross-dataset generalization, we evaluate on HO-3D and the out-of-domain object split of AffordPose under zero-shot settings, consistent with prior work~\cite{controlnet, Wu2024FastGraspEG, jiang2021graspTTA, grabdataset}. We exclude AffordPose from training due to its missing MANO parameters and partial noise, which are incompatible with our MANO-based differentiable model. Quantitative analysis justifying this exclusion is provided in the Appendix.



\subsection{Evaluation Metrics}
\label{sec:evaluation_metrics}
Following established evaluation protocols~\cite{grabdataset,halo,controlnet,Wu2024FastGraspEG}, we assess generated grasping poses using four criteria: (1) physical plausibility, (2) stability, (3) pose diversity, and (4) semantic alignment with language specifications.

\noindent\textbf{Physical Plausibility Assessment.} We evaluate physical plausibility through two metrics: (1) hand-object mutual penetration volume calculated by voxelizing both meshes at $1mm^3$ resolution and measuring intersection regions, and (2) contact ratio, which measures the percentage of grasp poses maintaining persistent surface contact.

\noindent\textbf{Grasp Stability Assessment.}
Stability evaluation follows prior physics-based approaches~\cite{Wu2024FastGraspEG,jiang2021graspTTA,grabdataset} through simulated grasp executions. We quantify stability by measuring the gravitational displacement of the object’s center of mass, with lower displacement indicating greater robustness.

\noindent\textbf{Diversity Assessment.} Following diversity metrics from grasp generation literature~\cite{halo,Wu2024FastGraspEG,liu2023contactgen}, we apply K-means clustering (k=20) to 3D hand keypoints across all methods. Diversity is assessed via two measures: (1) entropy of cluster assignments, where higher values indicate more diverse distributions across clusters, and (2) average cluster size, reflecting grasp space coverage. While larger average cluster sizes reflect better coverage of the grasp space.

\noindent\textbf{Semantic Accuracy Assessment(ACC).} To evaluate semantic consistency, we categorize hand-object interactions into ten affordance classes like AffordPose~\cite{AffordPose}. A classifier trained during the data annotation phase is used to assess whether the generated hand poses align with the corresponding semantic affordance instructions. To validate the effectiveness of our classifier, we report its accuracy on unseen objects in the appendix.
\subsection{Grasp Generation Performance}
\label{experimen}
\noindent\textbf{In-Domain Evaluation.}
\label{sec:in-domain}
Our approach, which combines DAM and hierarchical spatial fusion, outperforms all competitors in the in-domain evaluation (see Tab.~\ref{tab:in-domain} and Fig.~\ref{fig:comparison-indomain}). It shows superior performance on the OakInk~\cite{oakink} and GRAB~\cite{grabdataset} datasets across all metrics: intrusion volume, simulation distance, grasp pose diversity, and semantic accuracy. These results highlight the effectiveness of our method in generating complex grasp poses, surpassing leading methods such as FastGrasp~\cite{Wu2024FastGraspEG}, D-VQVAE~\cite{D-vqvae}, and ControlNet~\cite{controlnet}.

\vspace{1mm}
\noindent \textbf{Out-of-Domain Evaluation.}
\label{sec:out-of-domain}
We further evaluated our model’s universal applicability on the HO-3D~\cite{ho3d} and AffordPose~\cite{AffordPose} datasets. As depicted in Tab.~\ref{tab:out-of-domain} and Fig.~\ref{fig:comparison-indomain}, out-of-domain evaluation validates our method’s outstandingly accurate semantic results, in addition to physical generalization and generation diversity. Our method hence emerges as an effective solution that can transcend domain boundaries.
\begin{figure}[t]
    \centering
    \includegraphics[width=0.4\textwidth]{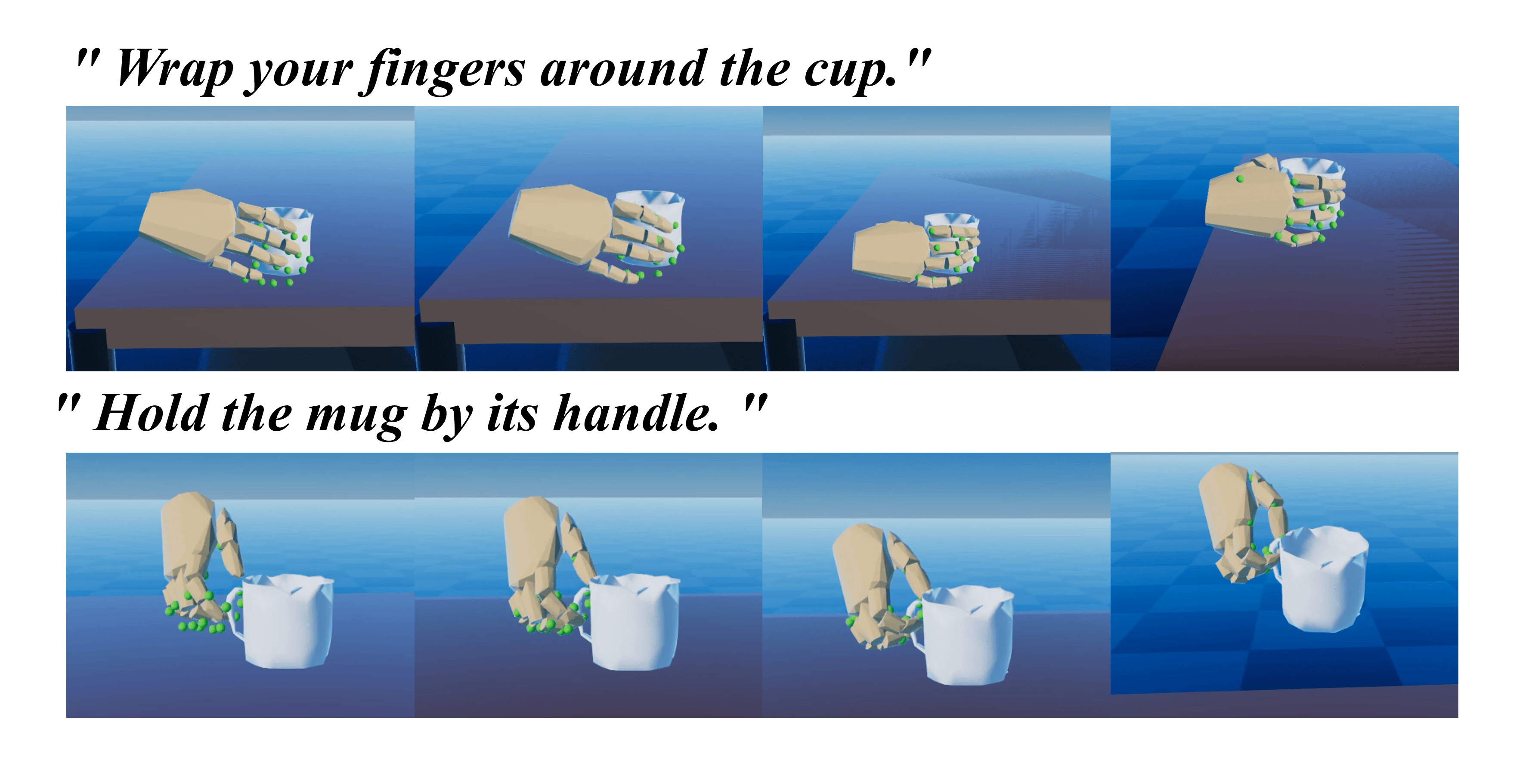}
    \caption{{Simulation results: grasping a single object under different instructions.} 
    }
    \vspace{-3ex}
    \label{fig:GraspProcess}
\end{figure}

\vspace{-2mm}
\subsection{Ablation Study}
We conduct a comprehensive ablation study to evaluate the impact of individual components on our framework’s performance. This analysis provides empirical evidence of each component's contribution, offering crucial context for subsequent experiments.

Tab.~\ref{tab:ablation-study} summarizes the key findings. Excluding object affordances results in a slight improvement in displacement distance but increases volume intrusion. This suggests that the model depends on object affordances to better capture spatial relationships and minimize hand-object collisions. Improved displacement occurs when object invasion causes immobilization, reducing movement. Thus, incorporating object affordances as cross-modal cues enhances the model's ability to capture spatial details of objects.

Removing the DAM module increases cluster sizes, as the diffusion model better captures global distributions, weakening the conditioning. In contrast, incorporating the DAM module results in a more concentrated output distribution, improving thematic coherence, adherence to physical constraints, and object contact rates. This novel use of the adaptive module transformation in Out-of-Distribution scenarios highlights robust generalization capabilities.

\begin{figure}[t]
    \centering
    \includegraphics[width=0.45\textwidth]{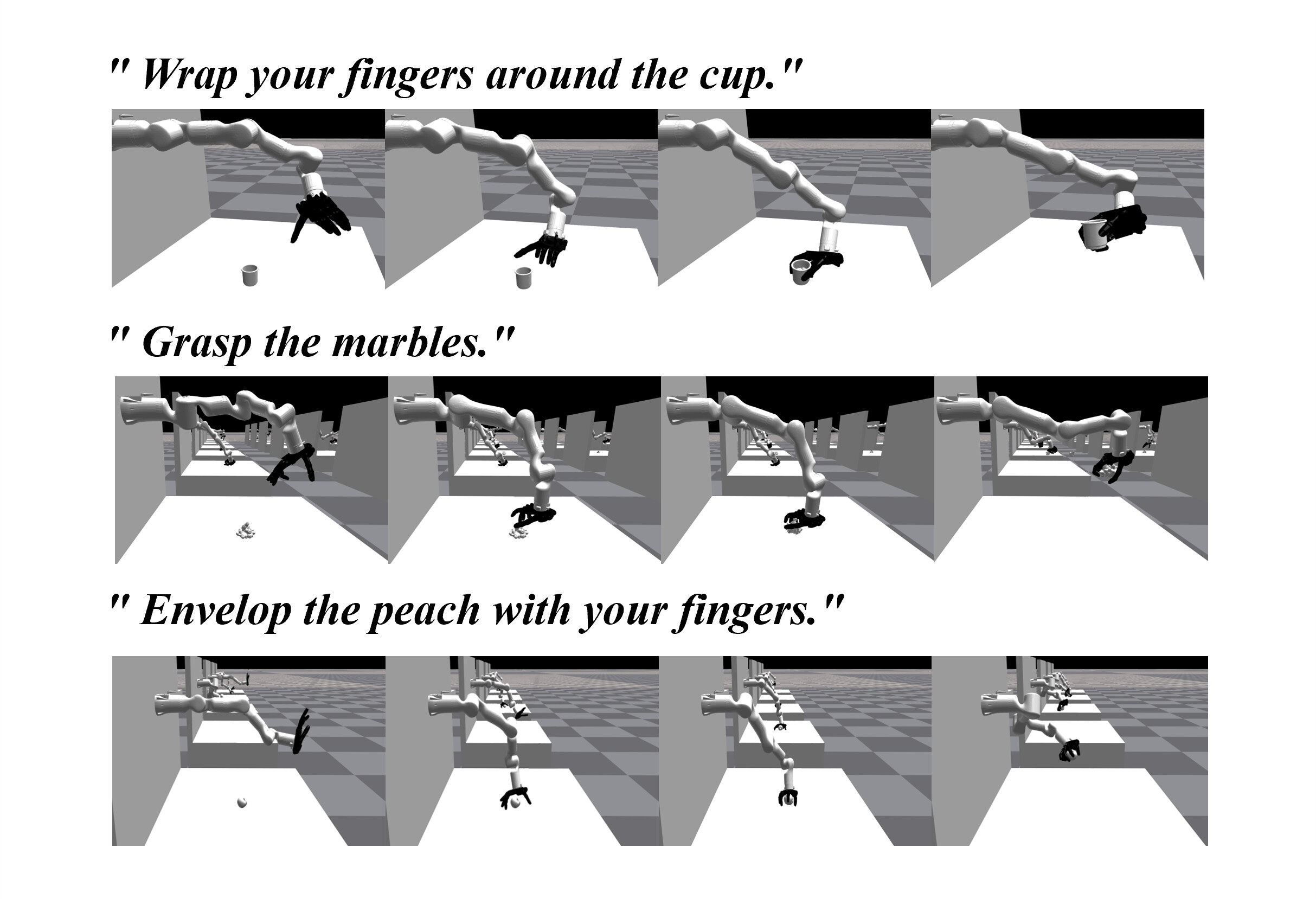}
    \caption{{Simulation results: grasping across multiple objects.}
    }
    \vspace{-3ex}
    \label{fig:GraspProcess2}
\end{figure}

\subsection{Downstream Applications}

To evaluate the downstream effectiveness of our generated grasps, we conducted experiments in the RaiSim~\cite{raisim} physics simulator. We employed both the D-Grasp~\cite{christen2022dgrasp} framework and the CrossDex~\cite{yuan2024cross} platform to execute dynamic grasp trajectories.

In our framework, AffordGrasp predicts a reference grasp pose $\bar{G}$ conditioned on the target object $O$ and the language instruction $L$. In D-Grasp, this reference is used to generate physically plausible and temporally consistent grasp motions. In CrossDex, the same reference grasp serves as a target goal configuration, guiding the policy to achieve it in simulation. During evaluation, each grasped object is lifted against gravity to assess stability. Fig.~\ref{fig:GraspProcess} and Fig.~\ref{fig:GraspProcess2} illustrate grasps of the same object under different instructions, as well as grasps of multiple objects.

%% file: sec/5_conclusion.tex
\section{Conclusion and Discussion}

This paper presents an automated annotation engine that enriches hand-object interaction datasets with linguistic instructions. By leveraging object affordance for cross-modal alignment and introducing a Distribution Adjustment Module, our method captures both spatial geometry and instruction semantics. It enhances physical plausibility through geometric grounding and improves alignment between language and 3D representation.

\vspace{-2mm}
\paragraph{Limitations.}
Our current framework primarily focuses on data-driven learning and does not explicitly incorporate physical priors such as gravity or friction. As illustrated in appendix Fig. 15, certain real-world effects may not be fully reflected in the generated results. Future work could benefit from integrating physics-based reasoning or simulation to further enhance grasp stability and realism.

\clearpage

\section{Acknowledgments}
This work was supported by NSFC 62350610269, Shanghai Frontiers Science Center of Human-centered Artificial Intelligence, and MoE Key Lab of Intelligent Perception and Human-Machine Collaboration (ShanghaiTech University). This work was also supported by HPC platform of ShanghaiTech University.

%% file: sec/X_suppl.tex
\clearpage
\FloatBarrier  
\setcounter{page}{1}
\twocolumn[  
\begin{center}
    {\Large \textbf{Supplementary Material}}
\end{center}
\vspace{1em}
]

\section{Overview of Material}

This supplementary material presents our detailed experiments, implementation, and additional visualizations. Sec.~\ref{autoencoder} investigates the impact of autoencoder architectures on hand representation. Sec.~\ref{sub:data engine} elaborates on the automatic data annotation engine, followed by the text generation pipeline in Sec.~\ref{text-notation}. In Sec.~\ref{sec:dataengine-affordance}, we present the object affordance prediction pipeline with experimental analysis. Sec.~\ref{sec:parameter_sen} and Sec.~\ref{sec:loss} discuss parameter sensitivity and training objectives. Additional affordance visualizations are provided in Sec.~\ref{sec:more_vis}. Sec.~\ref{sec:metric_acc} presents our evaluation scheme for assessing semantic consistency. Finally, Sec.~\ref{sec:simulation} and Sec.~\ref{sec:realrobot} demonstrate AffordGrasp’s execution in simulation and real-world experiments.

\section{Autoencoder Structure} \label{autoencoder}

We conduct a comparative analysis of various autoencoder architectures, as detailed in Tab.~\ref{tab:sub-ae}. Empirical results demonstrate that FastGrasp yields superior performance in terms of  reconstruction fidelity. Furthermore, FastGrasp achieves a significantly higher compression ratio compared to D-VQVAE~\cite{D-vqvae}, enabling more efficient latent representation. Consequently, we adopt FastGrasp as the backbone autoencoder for our framework.
\section{Inference Speed}

We investigated the integration of test-time adaptation~\cite{jiang2021graspTTA,liu2023contactgen} and gradient optimization~\cite{Wu2024AffordDPGD,Yang2024GuidanceWS} during the diffusion sampling process. However, we observed that both techniques incurred a substantial computational overhead, significantly compromising the real-time inference efficiency, as shown in Tab.~\ref{tab:sub_speed}. Therefore, we exclude them from our final inference pipeline to ensure efficiency.

\section{Data Engine}
\label{sub:data engine}

Our data engine leverages existing datasets—such as OakInk~\cite{oakink} and GRAB~\cite{grabdataset}—which primarily contain paired hand–object data. We aim to enrich these datasets with additional modalities through automated annotation, ultimately constructing a dataset that includes hand meshes, object point clouds, textual descriptions, and object affordance labels.

The data pipeline consists of two sequential modules with strict dependencies. The first module generates textual descriptions that characterize hand–object interactions, and the second module infers object affordances based on these generated descriptions. To simplify model training, we adopt an offline pipeline that directly outputs affordance labels. The workflow is strictly unidirectional: the second module is executed only after the successful completion of the first, preserving the dependency between text generation and affordance inference.

\begin{table}[t]
    \centering
    \setlength{\abovecaptionskip}{0pt} 
    \setlength{\belowcaptionskip}{0pt} 
    \resizebox{\linewidth}{!}{
    \begin{tabular}{c|cccc}
        \hline
        Method &
        \begin{tabular}{c} Simulation \\ Displacement $\downarrow$ \end{tabular} &
        \begin{tabular}{c} Penetration \\ Volume $\downarrow$ \end{tabular} &
        \begin{tabular}{c} Penetration \\ Distance $\downarrow$ \end{tabular} &
        \begin{tabular}{c} Contact \\ Ratio $\uparrow$ \end{tabular} \\
        \hline
        D-VQVAE~\cite{D-vqvae} & \textbf{1.61} & 0.94 & 5.17 & 98 \\
        \hline
        FastGrasp~\cite{Wu2024FastGraspEG} & 1.83 & \textbf{0.91} & \textbf{2.39} & \textbf{98} \\
        \hline
    \end{tabular}
    }
    \caption{Comparative experiments to evaluate the impact of different autoencoder structures.}
    \label{tab:sub-ae}
    \vspace{-2ex}
\end{table}

\begin{table}[t]
    \centering
    \setlength{\abovecaptionskip}{0pt} 
    \setlength{\belowcaptionskip}{0pt} 
    \resizebox{0.8\linewidth}{!}{
    \begin{tabular}{c|ccc} 
        \hline
        Method & TTA & Sample optimization & Ours \\ 
        \hline
        Time & 7.09s & 1.73s & 0.45s \\ 
        \hline
    \end{tabular}
    } 
    \caption{Different optimization method's inference efficiency.}
    \vspace{-3ex}
    \label{tab:sub_speed}
\end{table}



\subsection{Text Instruction Generation}
\label{text-notation}
\label{sec:dataengine-text}

\paragraph{Semantic Affordance Prediction.}
The AffordPose dataset~\cite{AffordPose} provides rich annotations including hand, object, and semantic affordance labels (e.g., \textit{Handle-grasp, No-grasp, Press, Lift, Wrap-grasp, Twist, Support, Pull, Lever, Null}). To leverage this, we design a classifier that takes the joint hand-object point cloud as input to predict the semantic affordance category corresponding to the grasping pose.

The model utilizes a pre-trained PointBERT~\cite{yu2021pointbert} backbone to process combined hand-object point clouds. The input pipeline consists of: 1) \textbf{Canonicalization:} Unifying the point clouds through coordinate system alignment; 2) \textbf{Sampling:} Downsampling to $N=4096$ points via Farthest Point Sampling (FPS) to ensure consistent input dimensions; and 3) \textbf{Feature Extraction:} Learning joint features through the transformer backbone.

For the 10-class affordance prediction task, we employ the standard cross-entropy loss:
\begin{equation}
\mathcal{L}_{cls} = -\sum_{i=1}^C y_i \log(p_i),
\end{equation}
where $C=10$ denotes the number of affordance categories, $y_i$ represents the ground-truth one-hot label, and $p_i$ is the predicted probability for class $i$.

We employ an initial classifier trained on AffordPose  to generate pseudo-labels for the unlabelled OakInk and GRAB datasets. The pipeline proceeds as follows: 
\begin{enumerate}
    \item  \textbf{High-confidence Pseudo-labeling:} We select reliable predictions by filtering samples based on the output probability distribution, adhering to a strict confidence interval (referencing the $3\sigma$ principle).
    \item \textbf{Human Validation:} To ensure label quality, we conduct manual verification on a subset of 100 randomly sampled instances.
    \item \noindent \textbf{Iterative Refinement:} We adopt a self-training paradigm, jointly training on the original AffordPose data and the validated pseudo-labels. This process is repeated across multiple cycles to progressively expand label coverage until the target datasets are fully annotated.
\end{enumerate}

\begin{figure}[t]
    \centering
    \includegraphics[width=0.4\textwidth]{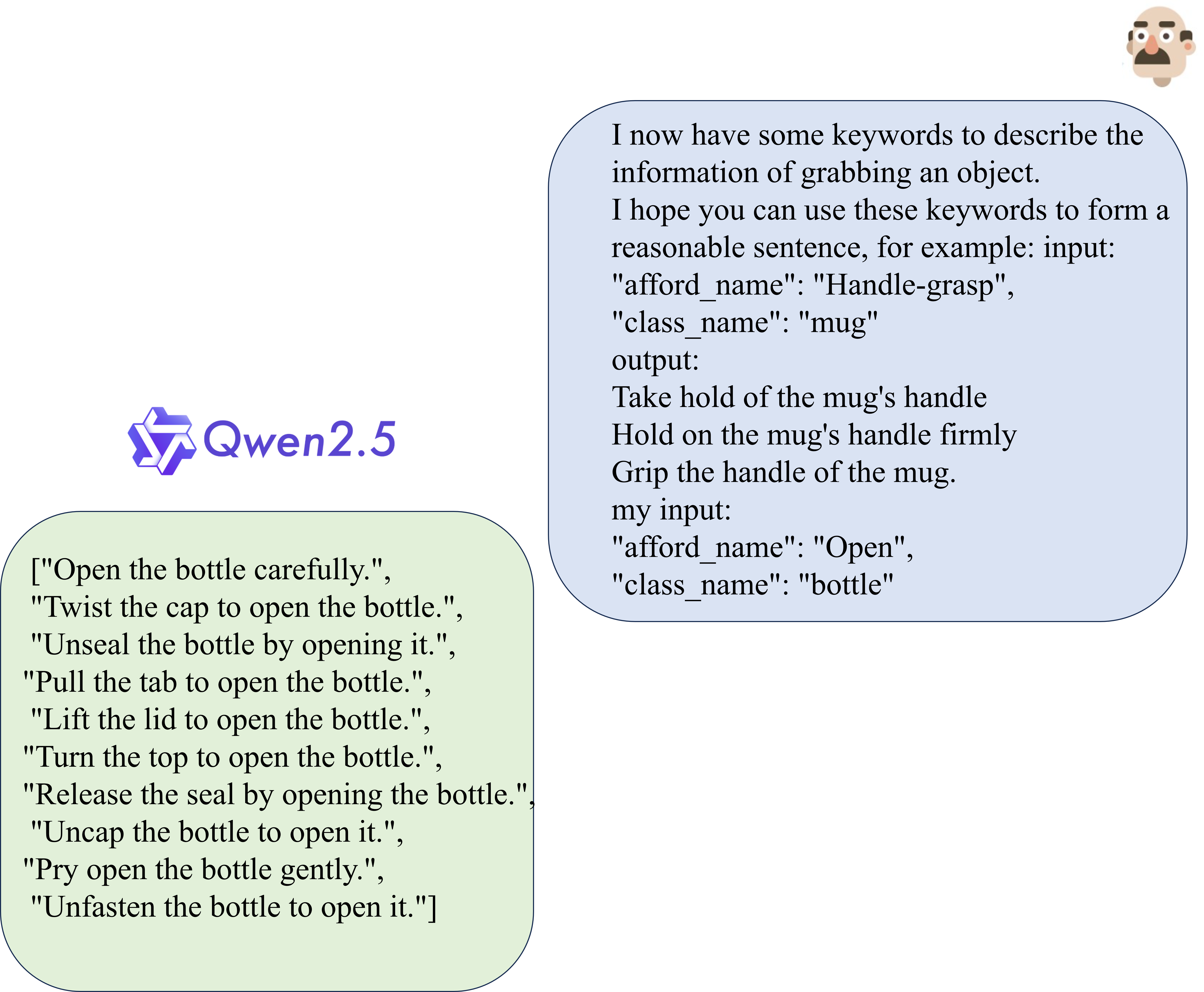}
    \caption{LLM generates the pipeline of language instructions.}
    \label{fig:sub_language}
    \vspace{-3ex}
\end{figure}

\paragraph{Instruction Generation.}
We generate language instructions based on the obtained language affordance and the class name of the object. We use Qwen (Fig.~\ref{fig:sub_language}) as our instruction generator, and generate the corresponding instructions through automated methods.

\begin{table}[t]
    \centering
    \setlength{\abovecaptionskip}{0pt}
    \setlength{\belowcaptionskip}{0pt}
    \resizebox{0.8\linewidth}{!}{
    \begin{tabular}{c|ccc}
        \hline
        Dataset & \begin{tabular}{c} train set \end{tabular} & \begin{tabular}{c} val set \end{tabular} & \begin{tabular}{c} test set \end{tabular} \\
        \hline
        Acc & 98.11\% & 97.90\% & 98.48\% \\
        \hline
    \end{tabular}
    }
    \caption{Classification accuracy experiment.}
    \label{tab:sub-classifier}
    \vspace{-2ex}
\end{table}

\paragraph{Experiment.}
To verify the validity of our data, we conducted experiments on the validation and test sets of the AffordPose dataset. As shown in Table \ref{tab:sub-classifier}, the model achieved an accuracy of 98\% on both the validation and test sets. This indicates that the model did not overfit the training set during the iterative optimization process.

\subsection{Object Affordance Prediction}
\label{sec:dataengine-affordance}

We generate object affordance representations using the annotated dataset to better align linguistic and geometric spatial embeddings. The model is trained on the AffordPose dataset, processing point clouds and language instructions to estimate per-point semantic adherence probabilities. This mapping associates textual semantics with localized operable regions in the point cloud through attention weights $p_{i,j} \in [0, 1]$, where $p_{i,j}$ indicates the focus priority for each spatial region during manipulation tasks. To address the class imbalance in affordance prediction, we optimize using a combined Focal Loss and Dice Loss objective:
\begin{figure}[t]
    \centering
    \includegraphics[width=0.5\textwidth]{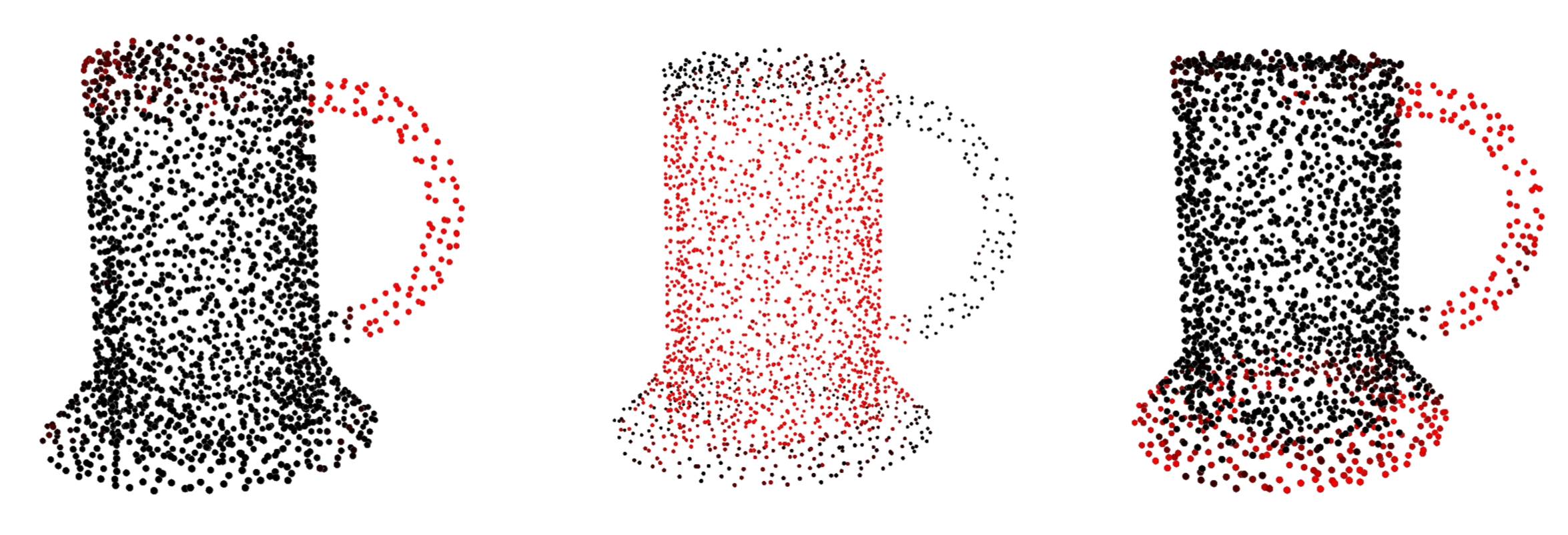}
    \centering
    \caption{\textbf{Object Affordance Visualization.}\\
    \label{fig:sub_aff_1}
    \textbf{Left.} Grip the handle of the mug.\\
    \textbf{Mid.} Wrap your hand around the mug.\\
    \textbf{Right.} Support the mug to prevent spills.}
    \vspace{-3ex}
\end{figure}
\begin{figure}[t]
    \centering
    \includegraphics[width=0.5\textwidth]{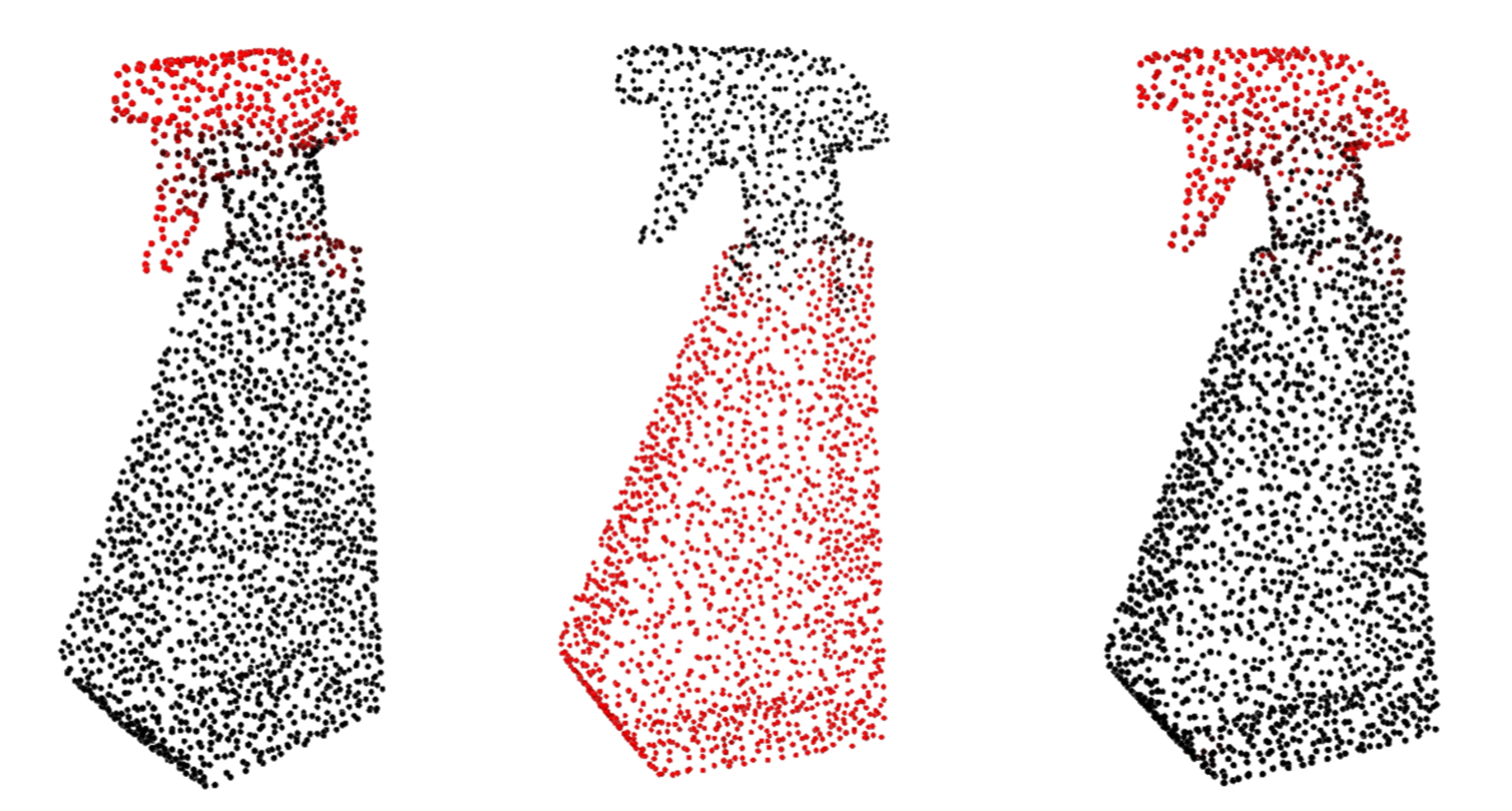}
    \centering
    \caption{\textbf{Object Affordance Visualization.}\\
    \label{fig:sub_aff_2}
    \textbf{Left.} Press the dispenser to avoid over-pouring.\\
    \textbf{Mid.} Wrap your fingers around the dispenser for a secure hold.\\
    \textbf{Right.} Support the bottle's handle to prevent spills.}
    \vspace{-3ex}
\end{figure}
\begin{figure}[t]
    \centering
    \includegraphics[width=0.5\textwidth]{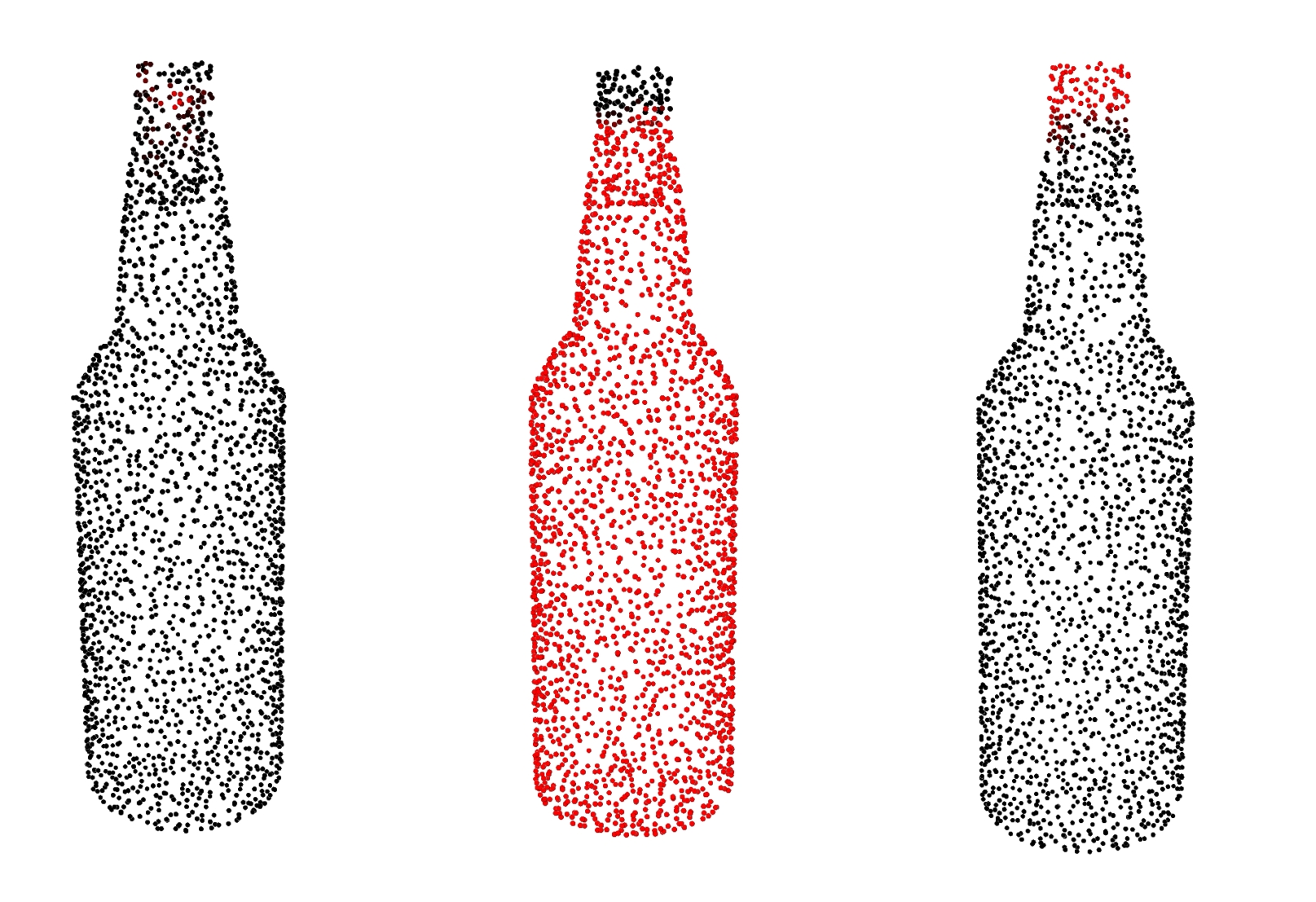}
    \centering
    \caption{\textbf{Object Affordance Visualization.}\\
    \label{fig:sub_aff_3}
    \textbf{Left.} Grip the handle of the bottle.\\
    \textbf{Mid.} Wrap-grasp the bottle to prevent spill.\\
    \textbf{Right.} twist the bottle to open it.}
\end{figure}

\begin{table*}[h]
\centering
\resizebox{\textwidth}{!}{
\begin{tabular}{|c|c|c|c|c|c|c|c|c|c|c|c|c|c|}
\hline
                  & bag & bottle & dispenser & earphone & faucet & handle-bottle & jar & keyboard & knife & laptop & mug & pot & scissors \\ \hline
IoU$\uparrow$               & 0.842 & 0.866 & 0.850 & 0.903 & 0.867 & 0.867 & 0.822 & 0.764 & 0.899 & 0.751 & 0.905 & 0.836 & 0.905 \\ \hline
AUC$\uparrow$               & 0.999 & 0.988 & 0.996 & 1.000 & 0.999 & 0.998 & 0.991 & 0.986 & 1.000 & 0.996 & 0.999 & 0.997 & 0.999 \\ \hline
SIM$\uparrow$               & 0.943 & 0.953 & 0.949 & 0.986 & 0.965 & 0.965 & 0.924 & 0.885 & 0.982 & 0.876 & 0.979 & 0.926 & 0.982 \\ \hline
MAE$\downarrow$               & 0.009 & 0.025 & 0.018 & 0.006 & 0.010 & 0.012 & 0.038 & 0.061 & 0.008 & 0.019 & 0.009 & 0.017 & 0.015 \\ \hline
\end{tabular}
}
\caption{Assessment for different object categories.}
\label{tab:sub_affo-1}
\end{table*}

\begin{table*}[h]
\centering
\small
\begin{tabular}{|c|c|c|c|c|c|c|c|c|c|c|}
\hline
 & Handle-grasp & Press & Lift & Wrap-grasp & Twist & Support & Pull & Lever & OVERALL \\ \hline
IoU$\uparrow$ & 0.889 & 0.786 & 0.872 & 0.910 & 0.807 & 0.723 & 0.717 & 0.870 & 0.855 \\ \hline
AUC$\uparrow$ & 1.000 & 0.993 & 1.000 & 0.991 & 0.998 & 0.990 & 0.999 & 0.999 & 0.996 \\ \hline
SIM$\uparrow$ & 0.975 & 0.903 & 0.965 & 0.975 & 0.923 & 0.851 & 0.836 & 0.967 & 0.948 \\ \hline
MAE$\downarrow$ & 0.009 & 0.032 & 0.001 & 0.030 & 0.013 & 0.029 & 0.006 & 0.009 & 0.019 \\ \hline
\end{tabular}
\caption{Assessment for different action.}
\label{tab:sub_affo-2}
\end{table*}

\begin{equation}
\begin{aligned}
\mathcal{L}_{\text{focal}}
&= -\frac{1}{N_{\text{points}}}
\sum_{i=1}^{N} \sum_{j}
\Big[
    \alpha\,(1 - p_{i,j})^{\gamma}\, g_{i,j} \log(p_{i,j})
    \\& + (1 - \alpha)\, p_{i,j}^{\gamma}\, (1 - g_{i,j}) \log(1 - p_{i,j})
\Big],
\end{aligned}
\label{eq:LOSS-focal}
\end{equation}

\begin{equation}
\begin{aligned}
\mathcal{L}_{\text{dice}}
&= \frac{1}{N} \sum_{i=1}^{N} \Big( 2 - \text{Dice}_{\text{pos}}^{(i)} - \text{Dice}_{\text{neg}}^{(i)} \Big),
\end{aligned}
\label{eq:LOSS-dice}
\end{equation}

\begin{equation}
\text{Dice}_{\text{pos}}^{(i)}
= \frac{2 \sum_{j} p_{i,j} g_{i,j}}{\sum_{j} p_{i,j} + \sum_{j} g_{i,j} + \epsilon},
\end{equation}

\begin{equation}
\text{Dice}_{\text{neg}}^{(i)}
= \frac{2 \sum_{j} (1 - p_{i,j})(1 - g_{i,j})}{\sum_{j} (1 - p_{i,j}) + \sum_{j} (1 - g_{i,j}) + \epsilon},
\end{equation}

\begin{equation}
\mathcal{L} = \mathcal{L}_{\text{focal}} + \lambda\mathcal{L}_{\text{dice}},
\end{equation}
where $\lambda$ is a balancing hyperparameter, $N$ is the batch size, $N_{\text{points}}$ is the total number of points in the batch, $p_{i,j}$ is the predicted probability for point $j$ in sample $i$, and $g_{i,j} \in \{0, 1\}$ is its ground-truth. For $\mathcal{L}_{\text{focal}}$, $\alpha$ and $\gamma$ are the standard balancing factor and focusing parameters. For $\mathcal{L}_{\text{dice}}$, we use a symmetric formulation by averaging the Dice coefficients for positive ($\text{Dice}_{\text{pos}}$) and negative ($\text{Dice}_{\text{neg}}$) classes, stabilizing training for imbalanced cases. $\epsilon$ is a small constant for numerical stability.

We validate our affordance prediction model through comprehensive benchmarking against state-of-the-art 3D affordance learning methods~\cite{laso,Azuma2021ScanQA3Q}, with detailed results in Tab.~\ref{tab:sub_affo-1} and Tab.~\ref{tab:sub_affo-2}. Our evaluation employs four established metrics: Area Under the Curve (AUC), Mean Intersection Over Union (mIoU), Similarity (SIM), and Mean Absolute Error (MAE), ensuring rigorous comparison across critical performance dimensions.

\noindent \textbf{AUC (Area Under the Curve):} In evaluation, AUC measures how effectively the model distinguishes affordance from non-affordance regions within objects. By analyzing classification performance across threshold variations, this metric captures the model's capacity to identify functionally relevant object parts under diverse conditions.

\noindent \textbf{mIoU (Mean Intersection Over Union):} This segmentation metric evaluates spatial alignment between predictions and ground truth masks. Computed as the mean IoU across all test samples, mIoU provides a comprehensive measure of segmentation accuracy by quantifying the overlap ratio between predicted and actual regions of interest.

\noindent \textbf{SIM (Similarity):} This metric quantifies the alignment between the model’s segmentation and the ground truth affordance region specified in the question, measuring the model’s ability to interpret textual queries and localize corresponding spatial regions. It is computed as:

\begin{equation}
\mathrm{SIM}(Y,M) = \sum_{i=1}^n \min(Y_i, M_i),
\end{equation}
\begin{equation}
 \quad \sum_{i=1}^n Y_i = \sum_{i=1}^n M_i = 1,
\end{equation}
where $Y$ and $M$  represent the ground truth and predicted segmentation masks, respectively, and $n$ is the total number of segmentation points (pixels). Both masks are normalized to form probability distributions over the spatial domain.
\begin{table*}[h]
\centering
\resizebox{\textwidth}{!}{
\begin{tabular}{cccccccc}
\hline
Dataset &
Head nums &
\multicolumn{1}{c}{\begin{tabular}{c} Penetration \\ Volume $\downarrow$ \end{tabular}} &
\multicolumn{1}{c}{\begin{tabular}{c} Simulation \\ Displacement $\downarrow$ \end{tabular}} &
\multicolumn{1}{c}{\begin{tabular}{c} Contact \\ Ratio $\uparrow$ \end{tabular}} &
\multicolumn{1}{c}{\begin{tabular}{c} Entropy $\uparrow$ \end{tabular}} &
\multicolumn{1}{c}{\begin{tabular}{c} Cluster \\ Size $\uparrow$ \end{tabular}} &
\multicolumn{1}{c}{\begin{tabular}{c} ACC $\uparrow$ \end{tabular}} \\
\hline
 & 1 & 8.32 & 1.66 & 96 & 2.79 & \textbf{4.14} & 76.33\% \\
OakInk~\cite{oakink}& 2 & 7.79 & \textbf{1.38} & 97 & 2.88 & 3.66 & 77.21\% \\
 & 4 & \textbf{7.32} & 1.43 & \textbf{98} & \textbf{2.94} & 3.74 & \textbf{80.08\%} \\
\midrule
 & 1 & 5.71 & 1.25 & 98 & 2.79 & \textbf{3.77} & \textbf{62.50\%} \\
GRAB~\cite{grabdataset} & 2 & 5.11 & \textbf{1.13} & \textbf{100} & 2.87 & 3.75 & \textbf{62.50\%} \\
 & 4 & \textbf{3.06} & 1.66 & 94 & \textbf{2.91} & 3.53 & \textbf{62.50\%} \\
\midrule
 & 1 & 8.73 & 2.5 & 94 & 2.81 & \textbf{3.81} & 70.00\% \\
HO-3D~\cite{ho3d} & 2 & 10.99 & \textbf{1.92} & 96 & \textbf{2.87} & 3.66 & \textbf{70.00\%} \\
 & 4 & \textbf{7.38} & 2.33 & \textbf{97} & 2.85 & 3.70 & 72.00\% \\
\midrule
 & 1 & 19.77 & \textbf{2.41} & 96 & 2.92 & \textbf{4.19} & 63.83\% \\
AffordPose~\cite{AffordPose} & 2 & 25.31 & 2.71 & \textbf{98} & \textbf{2.92} & 3.96 & 63.78\% \\
 & 4 & \textbf{10.36} & 3.59 & 91 & \textbf{2.92} & 3.93 & \textbf{69.71\%} \\
\bottomrule
\end{tabular}
}
\caption{\textbf{Parameter sensitivity experiment.} We performed parameter sensitivity experiments(Tab.~\ref{tab:ablation-study}) using the same setting}
\label{sub:Parameter-sensitivity}
\end{table*}

\noindent\textbf{MAE (Mean Absolute Error):} MAE quantifies the total error magnitude between predicted and ground truth affordance segmentations, disregarding directional bias. This metric evaluates the model's pixel-level accuracy in segmenting object parts relevant to linguistic queries, measuring its ability to interpret affordance cues from natural language instructions:

\begin{align}
    \mathsf{MAE}(Y,M)=\sum_i^n|Y_i-M_i|,
\end{align}
where $n$ denotes the total number of points, and $Y$ and $M$ represent the ground truth and predicted segmentation masks respectively.

Visualization results in Figs.~\ref{fig:sub_aff_1},~\ref{fig:sub_aff_2} demonstrate our model's robustness. Notably, we intentionally introduced incorrect text instructions (Fig.~\ref{fig:sub_aff_3}, left) to test prediction consistency. Despite input discordance, the model adaptively suppresses spurious affordance predictions, exhibiting increased reliance on global point cloud features. This behavior suggests an inherent bias toward structural coherence over local text-instruction mismatches.

 


\section{Parameter Sensitivity Analysis}
\label{sec:parameter_sen}
We conduct sensitivity analysis on the cross-attention heads in our DAM module, evaluating how different configurations (1, 2, and 4 attention heads) impact grasp generation performance. As shown in Tab.~\ref{sub:Parameter-sensitivity}.

\begin{figure*}[ht]
    \centering
    \includegraphics[width=0.9\textwidth]{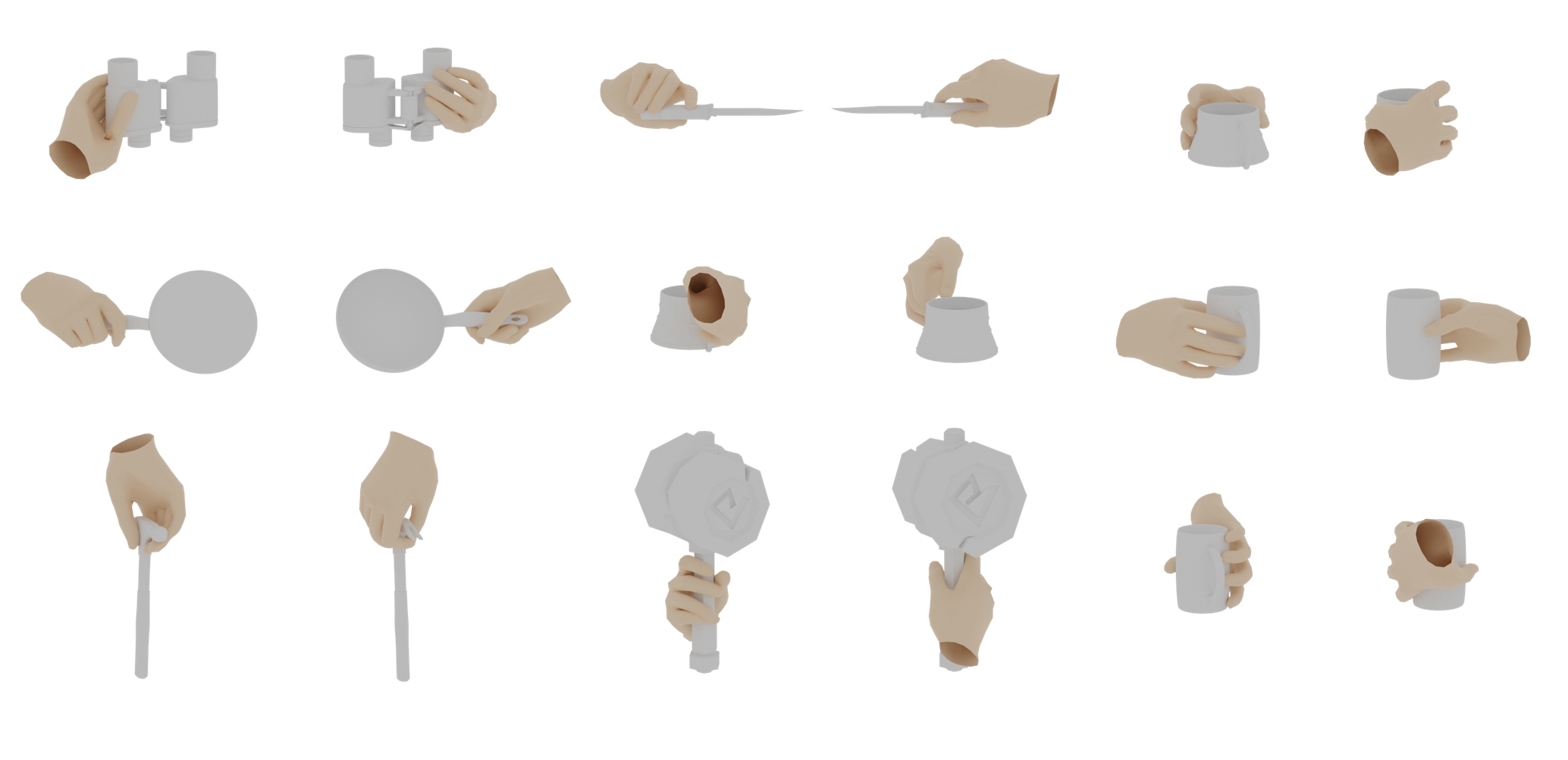}
    \caption{ Visualization of randomly selected grasping poses across different objects, each displayed from two viewpoints.}
    \label{fig:total1}
\end{figure*}
\begin{figure*}[ht]
    \centering
    \includegraphics[width=0.9\textwidth]{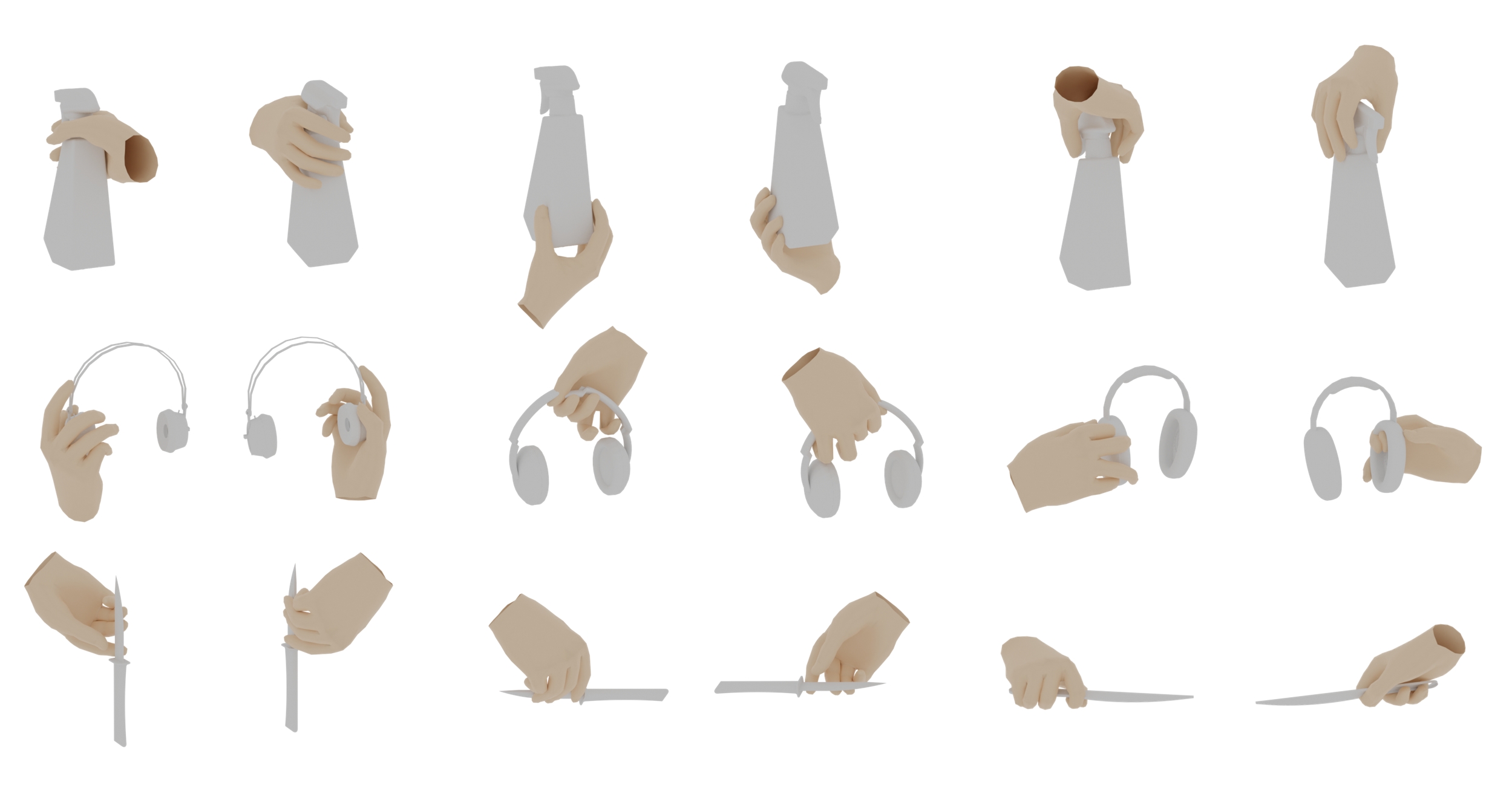}
    \caption{ Visualization of three grasping poses for the same object, each shown from two viewpoints.}
    \label{fig:total2}
\end{figure*}

\section{Loss function}
\label{sec:loss}
Based on Eq.~\ref{eq:diff}, we derive the original output distribution of the diffusion model through:
\begin{align}
\hat{h}_z = \frac{1}{\sqrt{\alpha_t}} (z^t - \sqrt{1-\alpha_t} \epsilon_{\theta}(z^t, f, t)),
\label{equ:approximation}
\end{align}
where $\alpha_t$ denotes the noise scheduling parameter. The training objective encompasses both the reconstruction loss and physical constraints as follows~\cite{Wu2024FastGraspEG,jiang2021graspTTA}:

\begin{equation}
h_p = Decoder(DAM(\hat{h}_z,f)),
\end{equation}
\begin{equation}
h_m = ManoLayer(h_p),
\end{equation}
\begin{equation}
\mathcal{L}_{recon} = \lambda_1 \mathcal{L}_{param} + \lambda_2\mathcal{L}_{mesh},
\end{equation}
\begin{equation}
\begin{aligned}
\mathcal{L}_{\text{mesh}}
&= 
\frac{1}{|h_v|} \sum_{x \in h_v} 
    \min_{y \in h_v^{\text{gt}}} \|x - y\|_2^2
\;\;
\\&+
\frac{1}{|h_v^{\text{gt}}|} \sum_{y \in h_v^{\text{gt}}} 
    \min_{x \in h_v} \|y - x\|_2^2 .
\end{aligned}
\label{eq:LOSS-mesh}
\end{equation}

\begin{equation}
\mathcal{L}_{param} = \text{MSE}(h_p, h_p^{\text{gt}}),
\label{eq:LOSS-param}
\end{equation}
where $\mathcal{L}_{param}$ indicates mean squared error loss between predicted $h_p$ and GT hand MANO parameters $h_p^{gt}$, $\mathcal{L}_{mesh}$ measures chamfer distance between the predicted hand vertices $h_v$ and the GT hand vertices $h_v^{gt}$, with $h_v$ derived from the hand mesh $h_m$.

To enforce physically plausible hand representations, we implement three additional constraint losses~\cite{Wu2024FastGraspEG,jiang2021graspTTA}:

\begin{equation}
\begin{aligned}
\mathcal{L}_{\text{consist}}
&= \mathrm{Consist}(h_m, h_m^{\text{gt}}, P_g) \\[4pt]
&= -\lambda_c
\frac{
    \sum_{i=1}^{B}
    \sum_{p=1}^{N_o}
    \mathbb{I}\!\left[d^{(i)}_{\text{pred}}(p) < \tau \right]
    \cdot
    \mathbb{I}\!\left[d^{(i)}_{\text{gt}}(p) < \tau \right]
}{
    \sum_{i=1}^{B}
    \sum_{p=1}^{N_o}
    \mathbb{I}\!\left[d^{(i)}_{\text{gt}}(p) < \tau \right]
},
\end{aligned}
\label{eq:LOSS-consistency}
\end{equation}

\begin{equation}
\begin{aligned}
\mathcal{L}_{\text{cmap}}
&= \mathrm{Contact}(h_m, P_g, M_c) \\[4pt]
&= \frac{1}{B} \sum_{i=1}^{B} 
    \min_{t \in \{1, \dots, T\}}
    \frac{
        \sum_{p=1}^{N_o} 
        M_c^{(i)}(p,t)
        \, d_{pred}^{(i)}(p)
    }{
        \sum_{p=1}^{N_o} 
        M_c^{(i)}(p,t)
    },
\end{aligned}
\label{eq:LOSS-cmap}
\end{equation}
\begin{equation}
\begin{aligned}
\mathcal{L}_{\text{penetr}}
&= \mathrm{Penetra}(h_m, P_g) \\[4pt]
&= \lambda_p
\frac{1}{B}
\sum_{i=1}^{B}
\sum_{p=1}^{N_o}
\mathbb{I}\!\left[
\mathrm{Inside}\!\left(P_g^{(i)}(p), h_m^{(i)}\right)
\right]
\cdot
d^{(i)}_{pred}(p).
\end{aligned}
\label{eq:LOSS-penetr}
\end{equation}

\begin{figure*}[ht]
    \centering
    \includegraphics[width=0.8\textwidth]{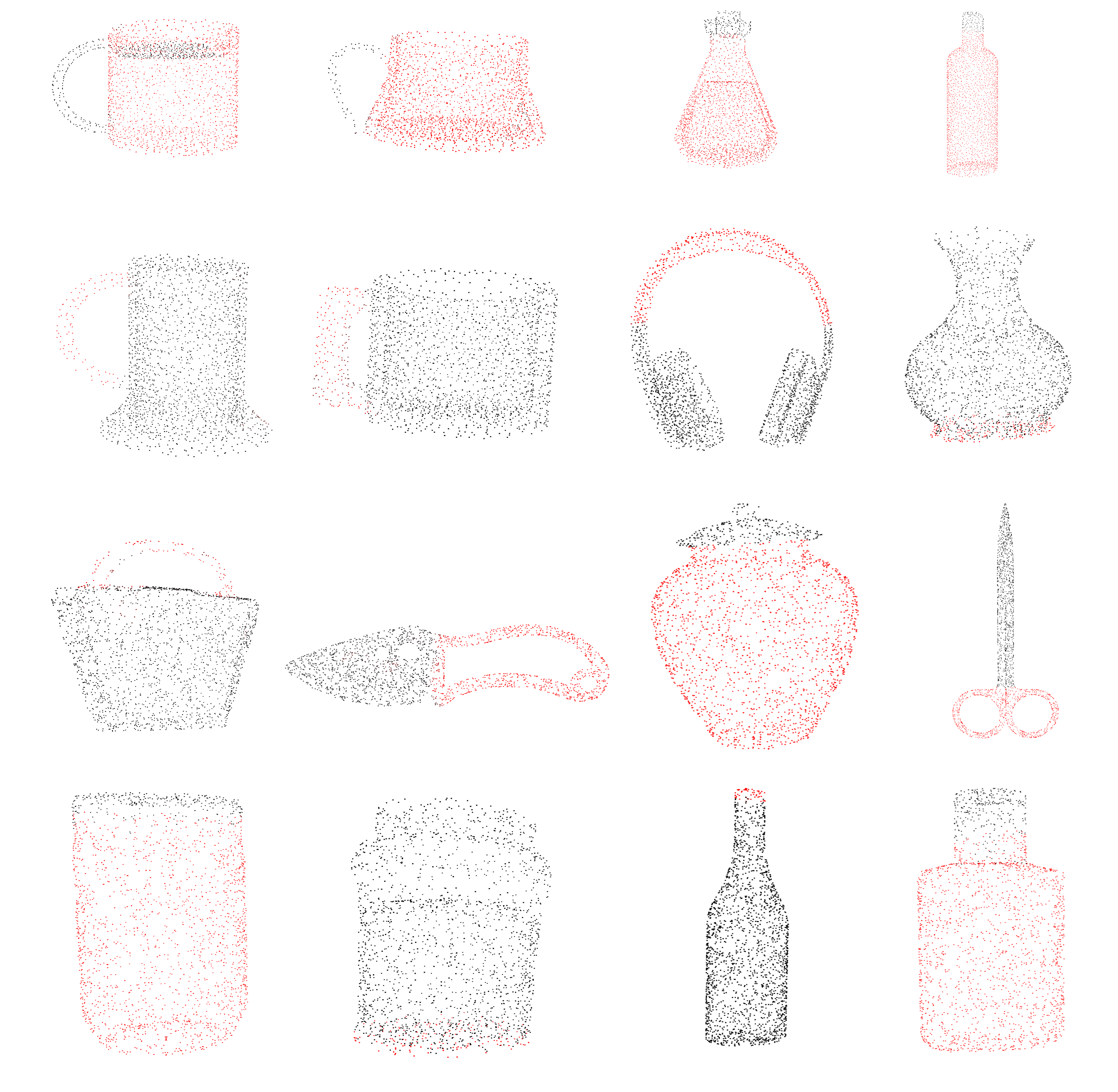}
    \caption{ Visualizations of affordance generation for multiple objects.}
    \label{fig:sub-aff2}
\end{figure*}

The predicted hand mesh’s contact region with the object mesh, denoted as $P_g$, which we aim to grasp, is kept consistent with the ground-truth (GT) hand mesh’s contact region through the consistency loss $\mathcal{L}_{\text{consist}}$ (Eq.~\ref{eq:LOSS-consistency}).
Specifically, $h_m$ and $h_m^{\text{gt}}$ represent the predicted and GT hand meshes, respectively, while $P_g$ denotes the object mesh. $B$ is the batch size, $N_o$ the number of object points, $\tau$ the contact distance threshold, and $\lambda_c$ a weighting coefficient. The distance term $d^{(i)}_{\text{pred}}(p)$ (or $d^{(i)}_{\text{gt}}(p)$) indicates the shortest distance from the $p$-th object point to the predicted (or GT) hand surface in the $i$-th sample.

The contact-map loss $\mathcal{L}_{\text{cmap}}$ (Eq.~\ref{eq:LOSS-cmap}) encourages the generated hand mesh $h_m$ to maintain meaningful contact with the object mesh $P_g$. Here, $M_c^{(i)}(p,t)$ denotes the binary contact mask for the $p$-th object point at the $t$-th contact frame, and $T$ is the total number of contact frames considered. 

The penetration loss $\mathcal{L}_{\text{penetr}}$ (Eq.~\ref{eq:LOSS-penetr}) serves as a physical constraint to penalize interpenetrations between the hand and object meshes. The indicator $\mathrm{Inside}(P_g^{(i)}(p), h_m^{(i)})$ checks whether the $p$-th object point lies inside the predicted hand mesh. The coefficient $\lambda_p$ balances the penalty term’s contribution. 

Our total loss function for training the DAM (Fig.~\ref{fig:DAM}) can be written as:
\begin{equation}
L=L_{recon}+\lambda_3L_{consist}+\lambda_4L_{cmap}+\lambda_5L_{penetr},
    \label{Loss: Total_loss}
\end{equation}
where the weight balancing coefficients $\lambda_{1}$, $\lambda_{2}$, $\lambda_{3}$, $\lambda_{4}$, $\lambda_{5}$ are used to balance these factors.

Through joint optimization of physical constraints and reconstruction objectives, our model effectively learns the physical interactions between hand meshes and objects, generating grasping poses that conform to natural physical principles. The two-stage training framework ultimately enables single-stage inference without requiring test-time adaptation (TTA), while still producing high-quality grasping configurations.

{
\begin{figure}[t]
    \centering
    \includegraphics[width=0.4\textwidth]{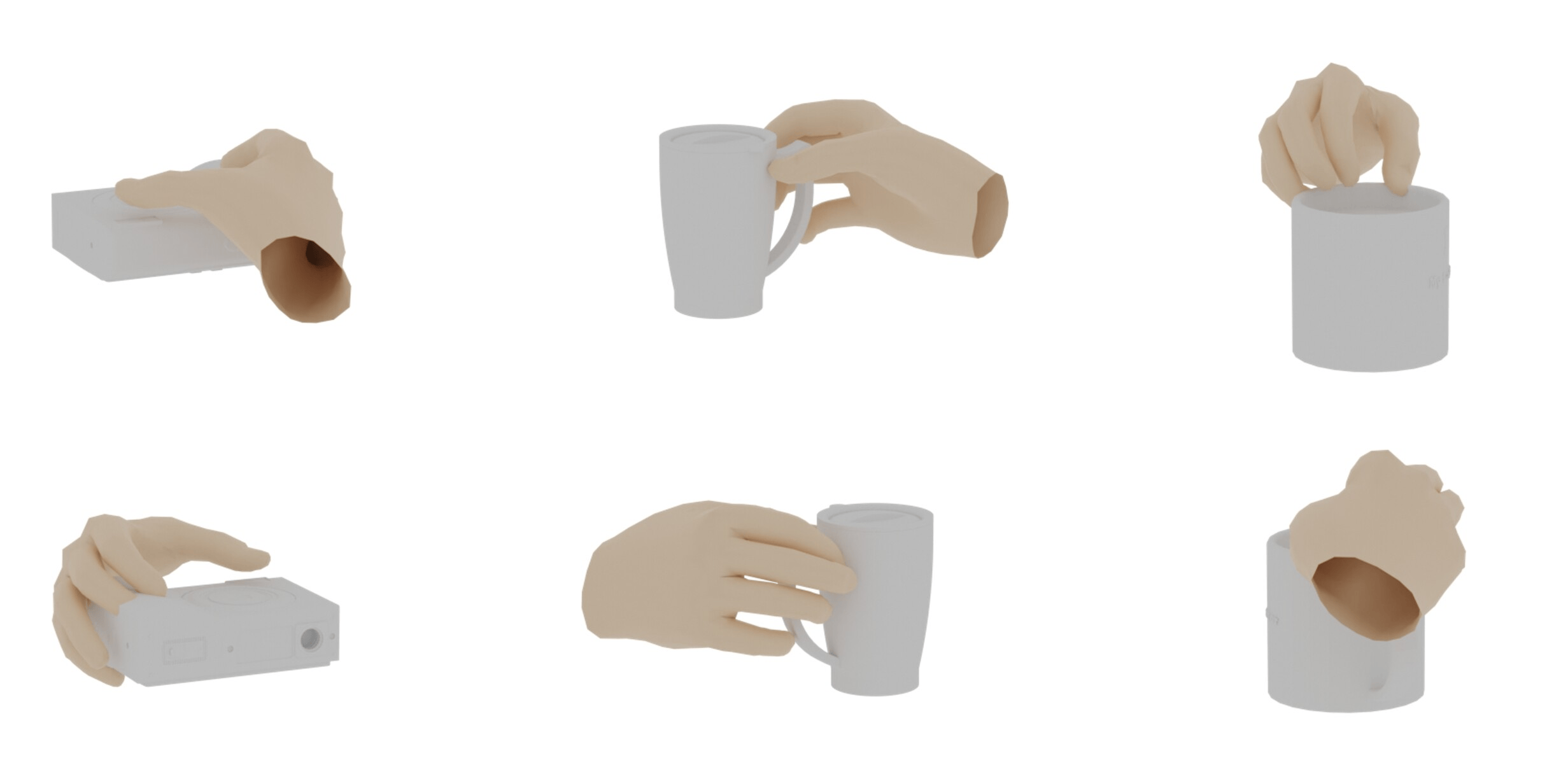}
    \caption{\textbf{Failure cases of our method.} Each pair displays a sample from two views.}
    \label{fig:limitation}
    \vspace{-3ex}
\end{figure}
}
\section{Visualization Result}
\label{sec:more_vis}
We provide additional visualizations in Fig.~\ref{fig:total1}, Fig.~\ref{fig:total2}, and Fig.~\ref{fig:sub-aff2}.
Fig.~\ref{fig:total1} presents randomly sampled grasping poses generated for different objects, demonstrating the generalization capability of our method.
Fig.~\ref{fig:total2} shows how varying textual prompts for the same object lead to distinct grasping configurations, indicating that textual guidance effectively modulates the generated poses.
Finally, Fig.~\ref{fig:sub-aff2} visualizes the outputs of our affordance generator, providing intuitive evidence of how object-level affordance cues guide grasp synthesis.

\section{Semantic Accuracy Assessment (ACC)}
\label{sec:metric_acc}

To evaluate the semantic consistency between the generated grasps and the input instructions, we adopt the taxonomy from AffordPose~\cite{AffordPose}, which defines ten distinct affordance classes: \{\textit{Handle-grasp, No-grasp, Press, Lift, Wrap-grasp, Twist, Support, Pull, Lever, Null}\}.

We utilize the semantic classifier (described in Sec.~\ref{sec:dataengine-text}) to assess whether the generated hand pose semantically matches the target object affordance. The prediction process is formally defined as:
\begin{equation}
\mathbf{p} = \text{classifier}(h_v, P_g),
\end{equation}
where classifier denotes the pre-trained classifier, $h_v$ represents the hand vertices, and $P_g$ is the object point cloud. The output $\mathbf{p} \in \mathbb{R}^{10}$ represents the confidence distribution over the ten affordance categories. The final predicted class $\hat{y}$ is determined by:
\begin{equation}
\hat{y} = \operatorname*{argmax}_{k} (\mathbf{p}),
\end{equation}
where $k$ denotes the affordance category index. We compute the Semantic Accuracy (ACC) as the percentage of samples for which the predicted class $\hat{y}$ matches the ground-truth text instruction.

\section{Simulation Experiment}
\label{sec:simulation}
To evaluate the performance of \textbf{AffordGrasp} in simulation, we conduct controlled experiments in a physics-based environment. Our pipeline first applies \textbf{CrossDex} to map MANO-based grasp predictions generated by AffordGrasp into the ShadowHand joint space. This remapping ensures kinematic feasibility and prevents implausible joint configurations.

The adapted ShadowHand grasps are then used as \emph{goal rewards} within the CrossDex reinforcement learning framework. During training, the policy is guided toward these target grasps, enabling AffordGrasp’s grasp priors to be executed as physically consistent behaviors in simulation.

\begin{figure}[t]
    \centering
    \includegraphics[width=0.5\textwidth]{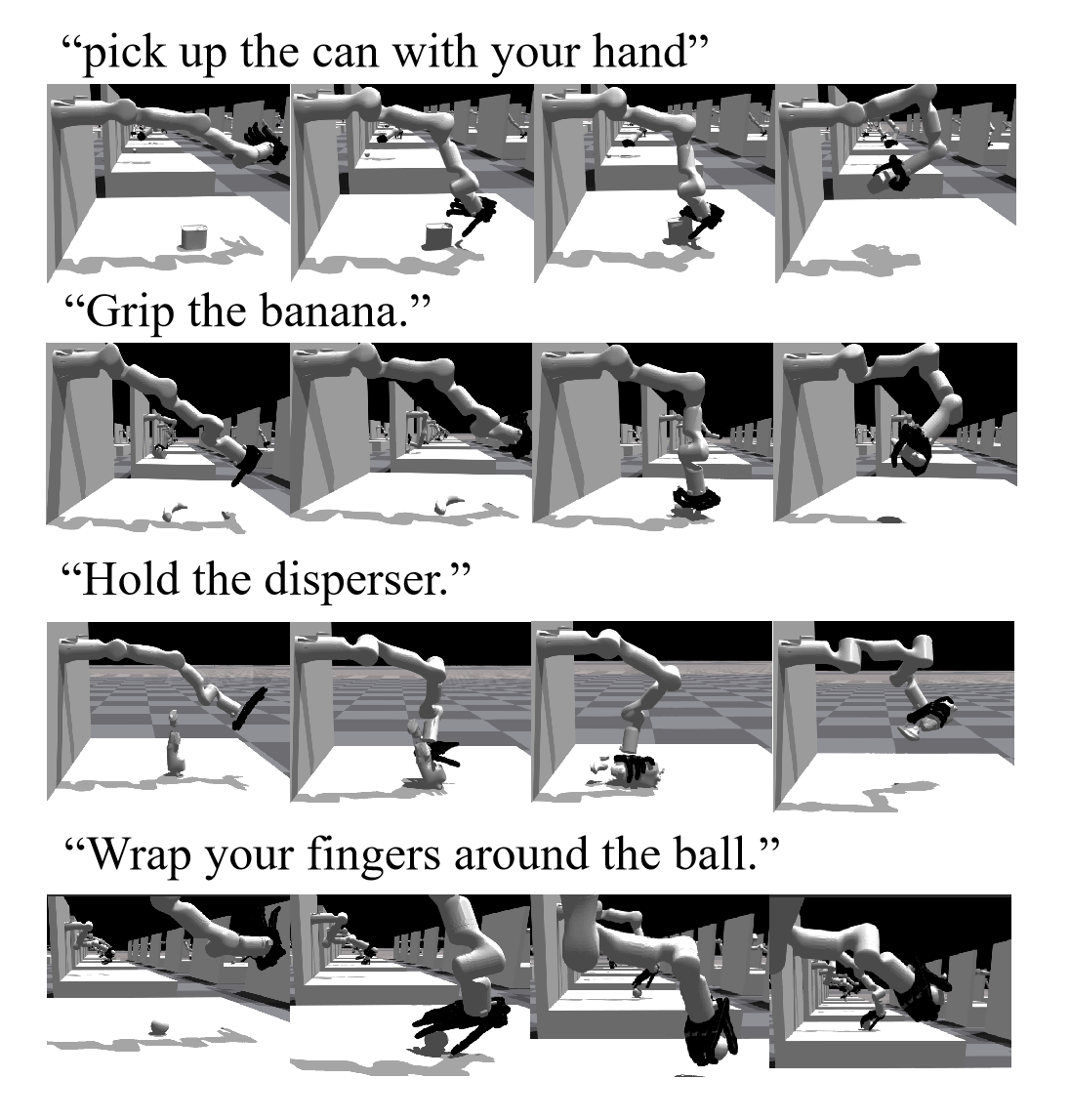}
    \vspace{-3ex}
    \caption{\textbf{Grasp execution results guided by AffordGrasp in the simulation environment.} 
    }
    \label{fig:simulation}
    \vspace{-3.5ex}
\end{figure}

We report grasp success rates, MPJPE, and FOL metrics, along with qualitative visualizations to demonstrate the effectiveness of our approach. As shown in Tab.~\ref{tab:simulation_results} and Fig.~\ref{fig:simulation}, AffordGrasp achieves a success rate comparable to CrossDex. Our method reaches an MPJPE of $0.161$ and an FOL of $0.235$, indicating that the RL policy produces grasp trajectories that are more physically consistent and stable under the guidance of the static reference. Visualizations further confirm that RL performs effective grasps with our affordance guidance.

\begin{table}[t]
\centering
\resizebox{\linewidth}{!}{
\begin{tabular}{lccc}
\toprule
Method & Success Rate (\%) & MPJPE $\downarrow$ & FOL $\downarrow$ \\
\midrule
CrossDex\cite{yuan2024cross}           &  92.90   &  -     &  0.260  \\
AffordGrasp(Ours)  &  \textbf{92.96}  &  0.161  &  \textbf{0.235} \\
\bottomrule

\end{tabular}
 }
\caption{Quantitative evaluation of AffordGrasp in the simulation environment.}
\vspace{-3ex}
\label{tab:simulation_results}
\end{table}

\begin{figure*}[t] 
    \centering
    
    \begin{subfigure}{\linewidth}
        \centering
        \includegraphics[width=0.8\linewidth]{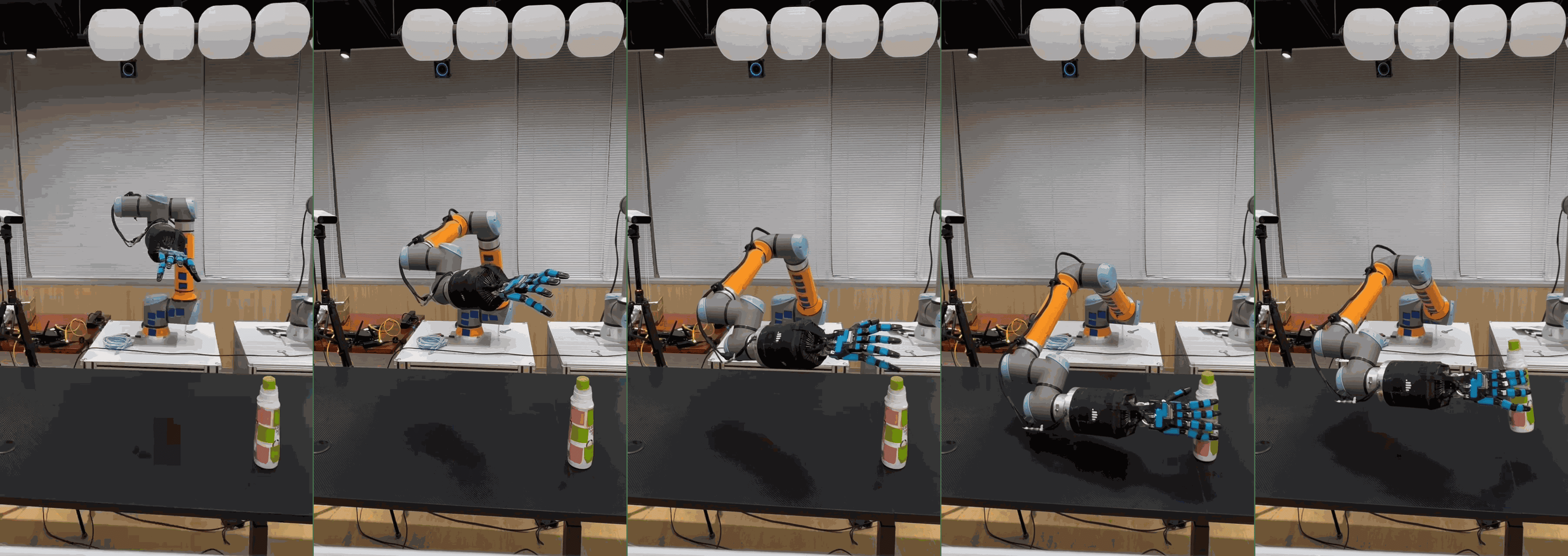} 
        \caption{Instruction: Wrap your hand around the bottle.} 
        \label{fig:sub1}
    \end{subfigure}

    \begin{subfigure}{\linewidth}
        \centering
        \includegraphics[width=0.8\linewidth]{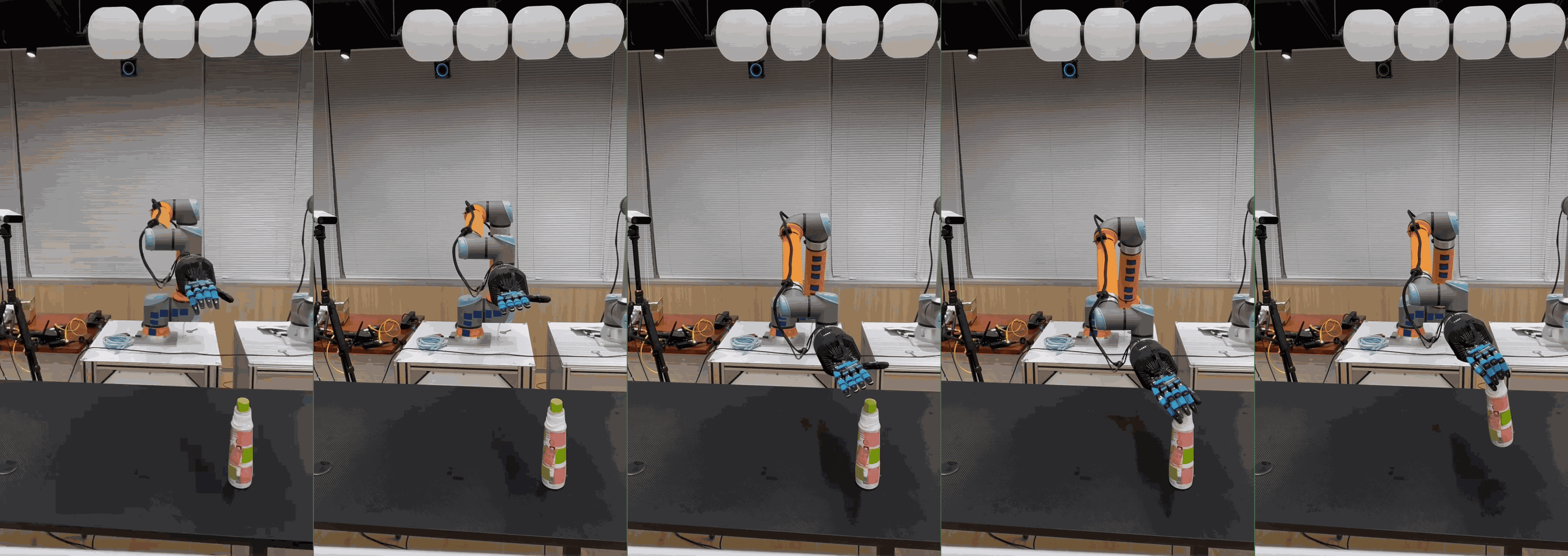} 
        \caption{Instruction: Twist the top of the bottle.}
        \label{fig:sub2}
    \end{subfigure}

    \begin{subfigure}{\linewidth}
        \centering
        \includegraphics[width=0.8\linewidth]{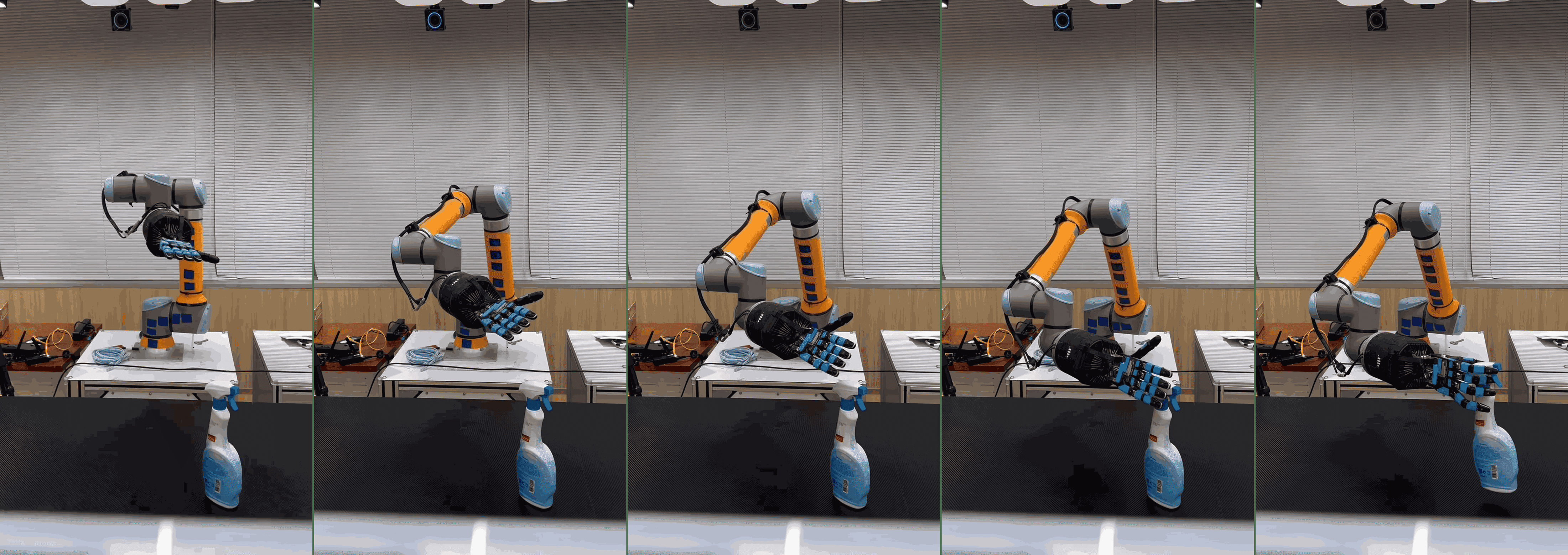} 
        \caption{Instruction: Press the dispenser to pour.}
        \label{fig:sub3}
    \end{subfigure}

    \begin{subfigure}{\linewidth}
        \centering
        \includegraphics[width=0.8\linewidth]{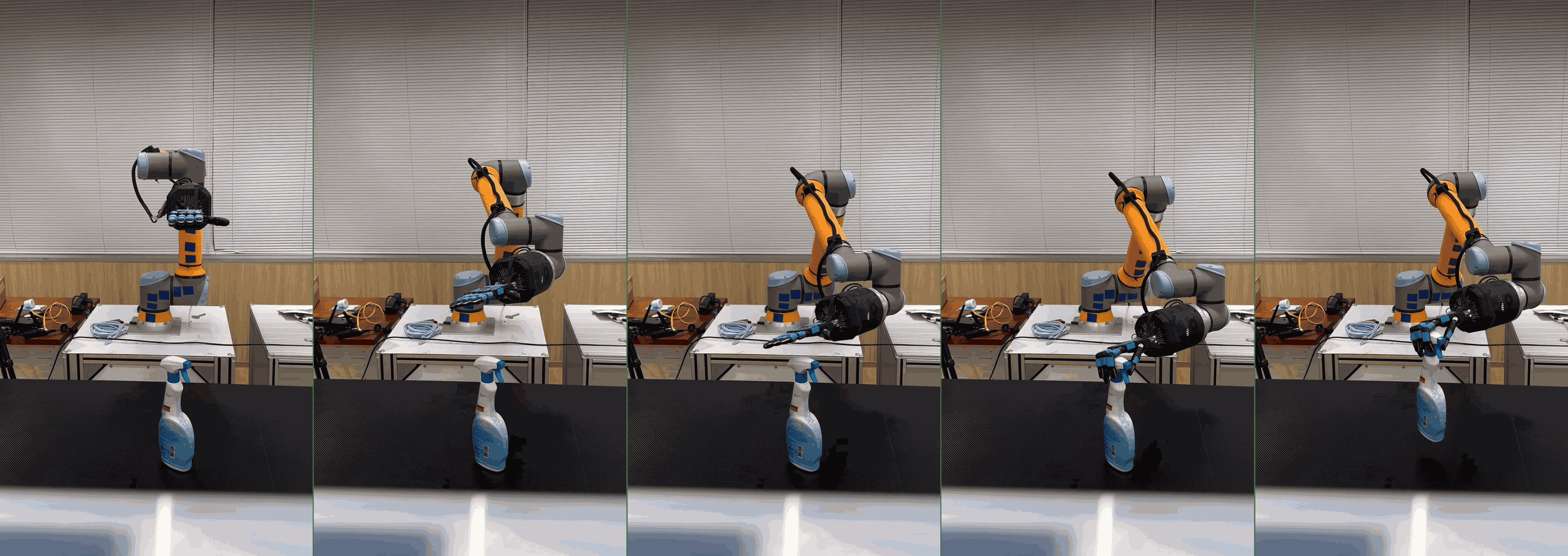} 
        \caption{Instruction: Twist the top of the dispenser to open it.}
        \label{fig:sub4}
    \end{subfigure}

    \caption{Real robot visualization.} 
    \label{fig:real_robot}
\end{figure*}
\section{Test on Real Robot}
\label{sec:realrobot}
We further evaluate AffordGrasp on a real robot. As shown in Fig.~\ref{fig:real_robot}, we execute the generated motions using the ShadowHand platform. For the same object, different language instructions lead to diverse execution outcomes. Throughout the entire process, the hand maintains behavior that is consistent with the input semantics, which meets the essential requirements for successful grasping.